%% file: main.tex
\documentclass[acmtog]{acmart}
\acmSubmissionID{683}
\usepackage{booktabs} %

\usepackage[inkscapearea=page]{svg}
\usepackage{adjustbox}

\input{defs}

\usepackage[ruled]{algorithm2e} %

\SetAlFnt{\small}
\SetAlCapFnt{\small}
\SetAlCapNameFnt{\small}
\SetAlCapHSkip{0pt}

\setcopyright{rightsretained}
\acmJournal{TOG}
\acmYear{2024} 
\acmVolume{43} 
\acmNumber{6} 
\acmArticle{194} 
\acmMonth{12}
\acmDOI{10.1145/3687953}

\usepackage{todonotes}
\usepackage{multirow}
\usepackage{stfloats} 
\usepackage{xspace}
\usepackage{makecell}
\usepackage{colortbl}

\definecolor{tabfist}{rgb}{0.36, 0.54, 0.66}
\definecolor{tabsecond}{rgb}{0.94, 0.97, 1.0}
\definecolor{tabthird}{rgb}{0.82, 0.1, 0.26}

\begin{document}

\title{Quark: Real-time, High-resolution, and General Neural View Synthesis}

\author{John Flynn}
\orcid{0009-0007-5858-6613}
\email{jflynn@google.com}
\authornote{Joint first authorship.}

\author{Michael Broxton}
\orcid{0009-0006-3701-2322}
\email{broxton@google.com}
\authornotemark[1]

\author{Lukas Murmann}
\orcid{0000-0001-5756-1819}
\email{lmurmann@google.com}
\authornotemark[1]

\author{Lucy Chai}
\orcid{0000-0002-4667-3097}
\email{lucyrchai@google.com}

\author{Matthew DuVall}
\orcid{0009-0001-0832-5606}
\email{matthewduvall@google.com}

\author{Clément Godard}
\orcid{0000-0002-1365-9571}
\email{cgodard@google.com}

\author{Kathryn Heal}
\orcid{0000-0003-1390-4589}
\email{kheal@google.com}

\author{Srinivas Kaza}
\orcid{0009-0000-9777-3496}
\email{srinivaskaza@google.com}

\author{Stephen Lombardi}
\orcid{0000-0003-3370-7216}
\email{salombardi@google.com}

\author{Xuan Luo}
\orcid{0009-0004-4605-5223}
\email{xuluo@google.com}

\author{Supreeth Achar}
\orcid{0009-0000-8863-7599}
\email{supreeth@google.com}

\author{Kira Prabhu}
\orcid{0009-0007-8381-5481}
\email{kmathias@google.com}

\author{Tiancheng Sun}
\orcid{0009-0005-8116-0405}
\email{tckevinsun@google.com}

\author{Lynn Tsai}
\orcid{0009-0007-3795-1386}
\email{lytsai@google.com}

\author{Ryan Overbeck}
\orcid{0009-0004-8485-1837}
\email{rover@google.com}

\affiliation{%
 \institution{Google}
 \streetaddress{1600 Amphitheatre Parkway}
 \city{Mountain View}
 \state{CA}
 \postcode{94043}
 \country{USA}}

\renewcommand\shortauthors{Flynn, J., Broxton, M., Murmann L., et al}

\begin{abstract}
We present a novel neural algorithm for performing high-quality, high-resolution, real-time novel view synthesis. From a sparse set of input RGB images or videos streams, our network both reconstructs the 3D scene and renders novel views at 1080p resolution at 30fps on an NVIDIA A100. Our feed-forward network generalizes across a wide variety of datasets and scenes and produces state-of-the-art quality for a real-time method. Our quality approaches, and in some cases surpasses, the quality of some of the top offline methods. In order to achieve these results we use a novel combination of several key concepts, and tie them together into a cohesive and effective algorithm. We build on previous works that represent the scene using semi-transparent layers and use an iterative learned render-and-refine approach to improve those layers. Instead of flat layers, our method reconstructs layered depth maps (LDMs) that efficiently represent scenes with complex depth and occlusions. The iterative update steps are embedded in a multi-scale, UNet-style architecture to perform as much compute as possible at reduced resolution. Within each update step, to better aggregate the information from multiple input views, we use a specialized Transformer-based network component. This allows the majority of the per-input image processing to be performed in the input image space, as opposed to layer space, further increasing efficiency. Finally, due to the real-time nature of our reconstruction and rendering, we dynamically create and discard the internal 3D geometry for each frame, generating the LDM for each view. Taken together, this produces a novel and effective algorithm for view synthesis. Through extensive evaluation, we demonstrate that we achieve state-of-the-art quality at real-time rates.

\end{abstract}

\ccsdesc[500]{Computing methodologies~Rendering}
\ccsdesc[500]{Computing methodologies~Reconstruction}
\ccsdesc[500]{Computing methodologies~Neural networks}

\begin{teaserfigure}
  \centering
  \includegraphics[width=0.99\textwidth]{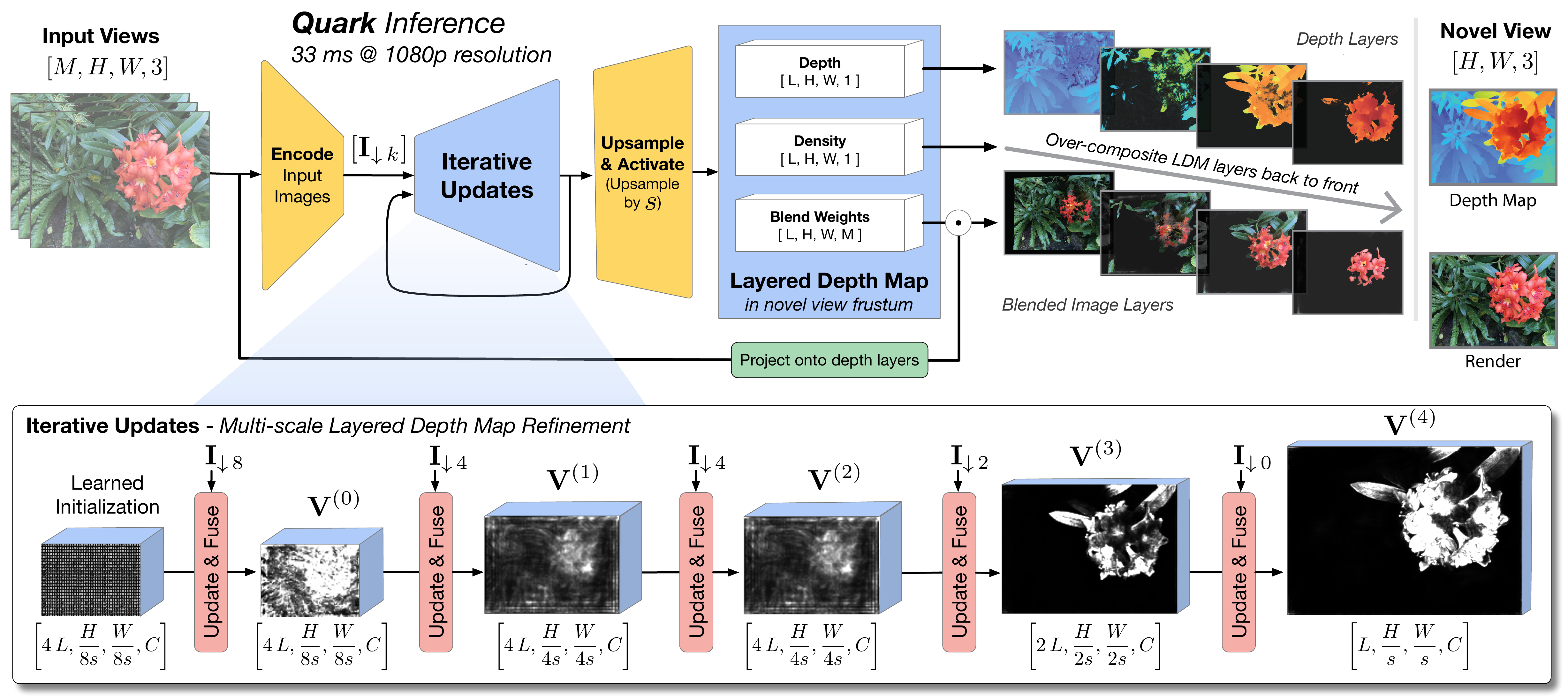}
  \vspace{-.1in}
  \caption{\textbf{Overview of the Quark Model}: Quark takes a set of multi-view images or video streams and reconstructs a compact layered depth map (LDM) representation that is used to perform image-based rendering. Network inference is fast; scene reconstruction and rendering \emph{combined} run at 30fps on a single NVIDIA A100 GPU at 1080p (1920$\times$1080) resolution, enabling Quark to perform high quality novel view synthesis \emph{on-demand} for a dynamic viewpoint even for scenes with moving content. During each time step the network creates a pyramid of downsampled and encoded input images using scalar downsampling factors $k$ and then infers an LDM in the frustum of the novel viewpoint through a series of $n$ \emph{Update \& Fuse} steps that use across-view attention to fuse information from multiple input views. This iterative, multi-scale approach also saves compute by gradually increasing spatial resolution while decreasing layer count. A final bilinear upsample by a scalar factor of $s$ followed by non-linear feature activation is used to produce an LDM at the final output resolution. See Section~\ref{sec:method} for more details on the notation and implementation.} 
  \label{fig:teaser}
\end{teaserfigure}

\keywords{Neural rendering, novel view synthesis, layered mesh representation, real-time feed-forward models}

\maketitle
\input{main_sections/01_intro}
\vspace{-.05in}
\input{main_sections/02_background}

\input{main_sections/03_method_v2}

\input{main_sections/05_results}

\input{main_sections/06_conclusion}

\begin{acks}
The authors would like to thank Peter Hedman and Ricardo Martin-Brualla for  thoughtful discussions and feedback during the course of this research. We are also grateful to the DL3DV~\cite{ling2023dl3dv} authors, especially Lu Ling, for their help in preparing evaluations using their dataset. We thank Alexander Familian for his help editing and doing the voice over for our submission video. And finally we would like to thank Jason Lawrence and Steve Seitz for their unfailing encouragement and support for this work.
\end{acks}

\bibliographystyle{ACM-Reference-Format}
\bibliography{references}

\begin{figure*}
    \centering
    \includegraphics[width=\linewidth]{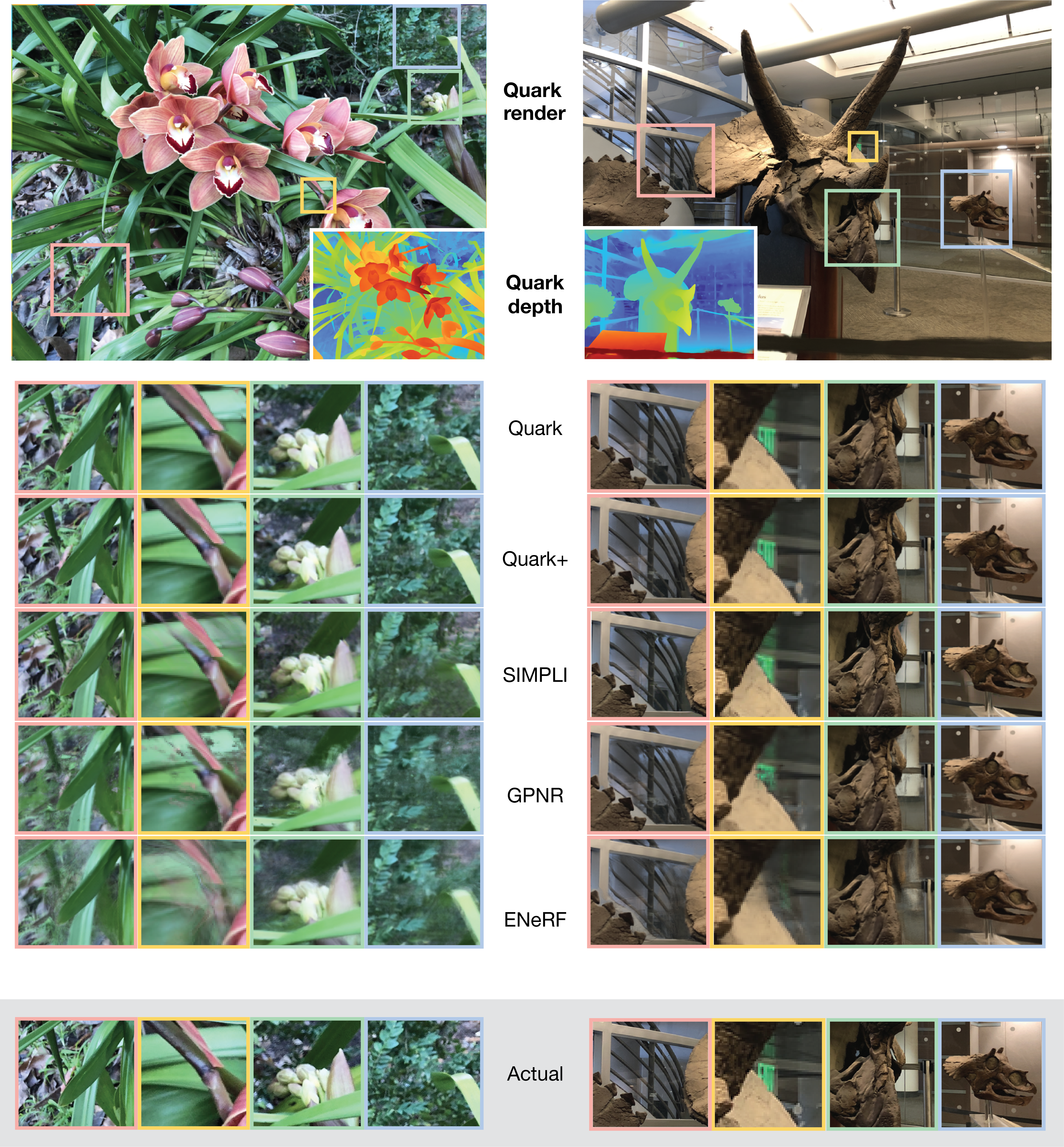}
    \caption{Comparison of Quark to current methods for generalizable neural view synthesis (for numerical results, see Table~\ref{table:comparison}). Quark preserves image details and preserves thin structures without blurring or image doubling.}
    \label{fig:renders-generalizable}
\end{figure*}

 \begin{figure*}
    \centering
    \includegraphics[width=\linewidth]{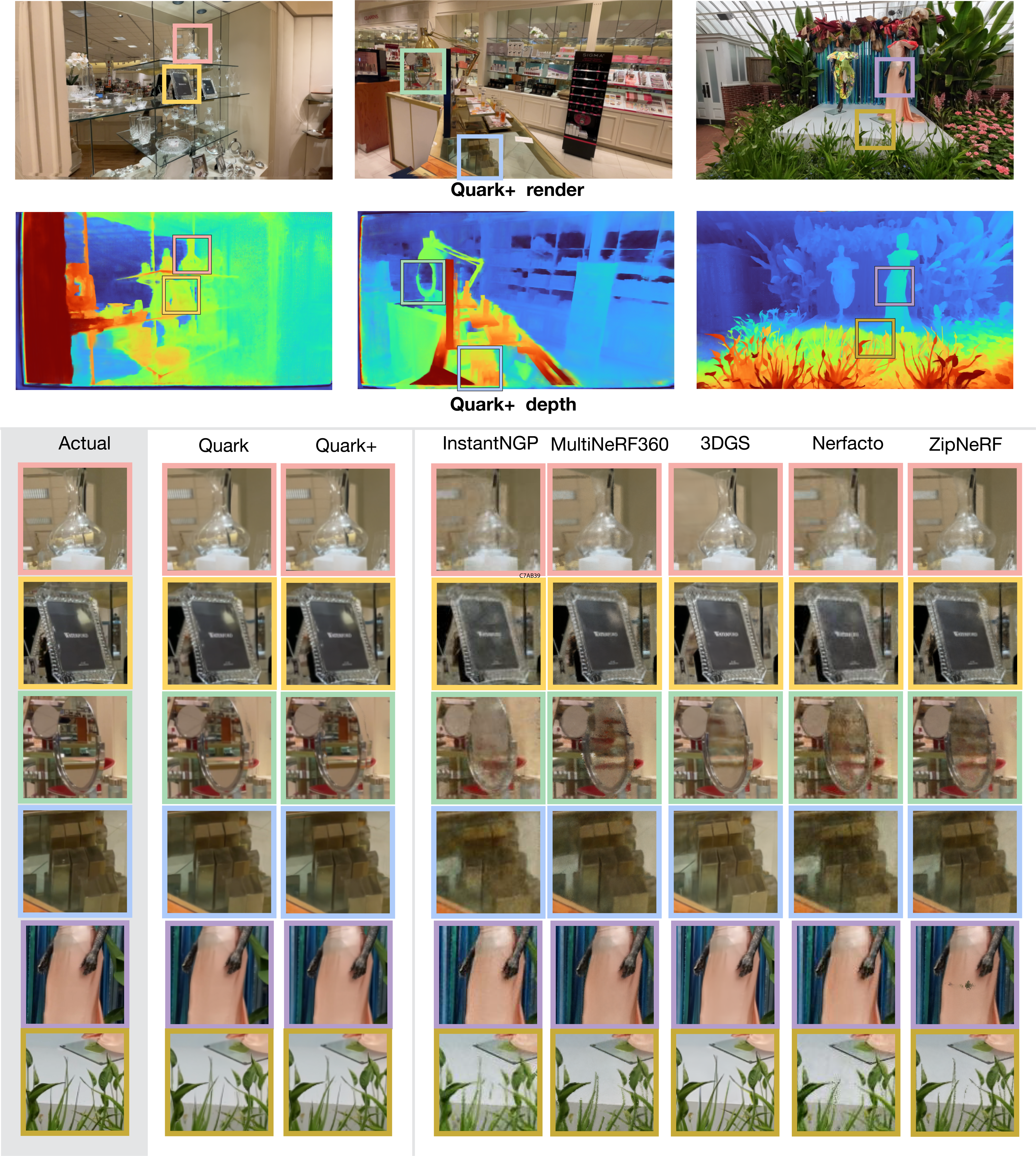}
    \caption{Comparisons of Quark to methods included in the DL3DV-10K NVS Benchmark (for numerical results, see Table~\ref{table:dl3dv}). Quark preserves crisp image detail and thin structures while faithfully rendering view-dependent effects like specular highlights and reflective surfaces.}
    \label{fig:renders-dl3dv}
\end{figure*}

\begin{figure*}
    \centering
    \includegraphics[width=\linewidth]{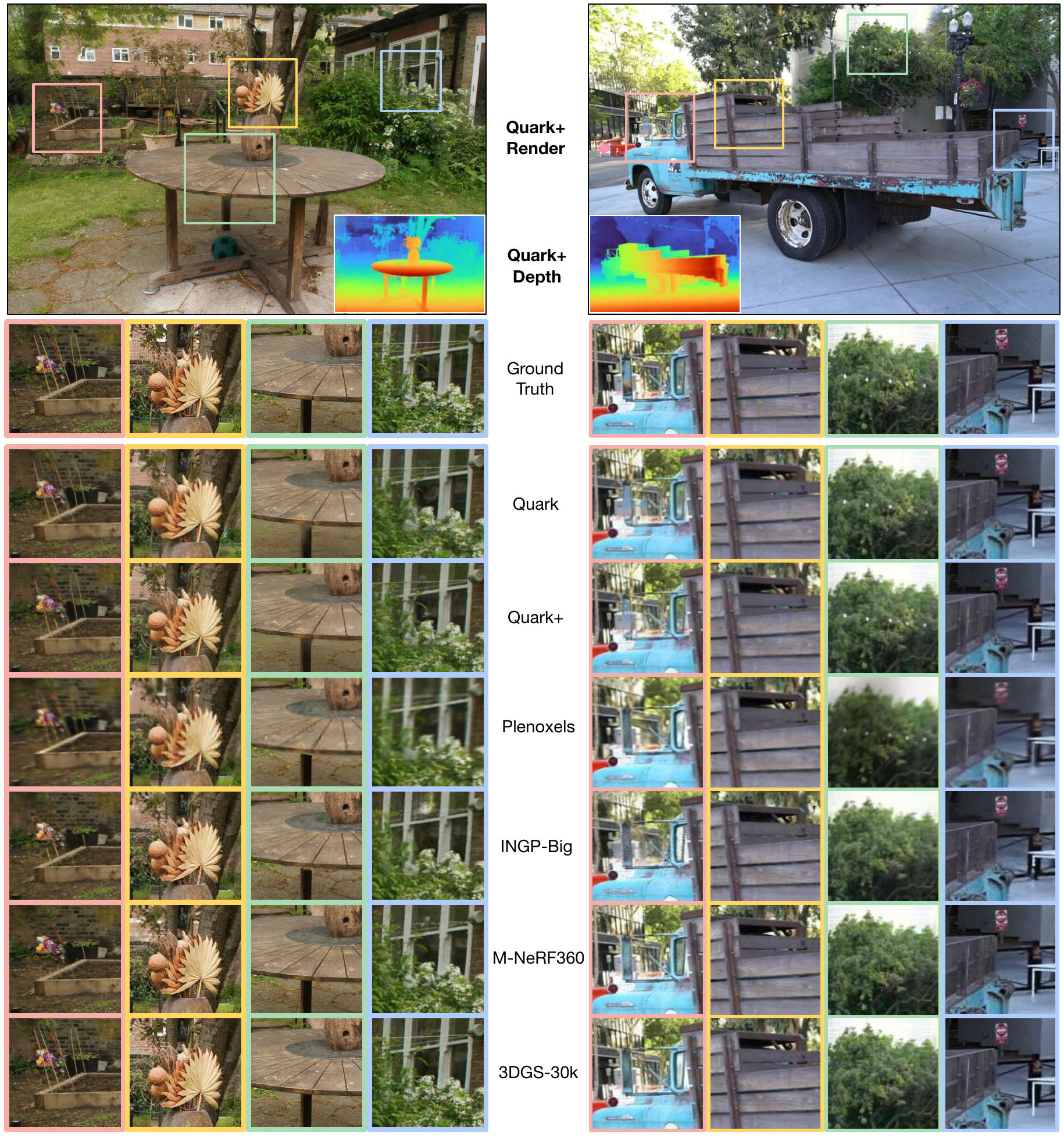}
    \caption{Comparisons of methods on scenes from the MipNeRF-360 and Tanks \& Temples datasets (for numerical results, see Table~\ref{tab:comparisons3dgs}). Quark closely matches the resolution of the target image and preserves thin structures as well as or better than other methods. }
    \label{fig:renders-3dgs}
\end{figure*}
\clearpage
\input{supp_sections/01_appendix}

\end{document}

%% file: defs.tex
\def\numimages{M}
\def\blendweights{\mathbf{\beta}}

\def\volume{\mathbf{V}}
\def\residual{\Delta}

\def\Norm{\mathscr{N}}

\newcommand{\cellg}{\cellcolor{green!20}}
\newcommand{\celly}{\cellcolor{yellow!40}}
\newcommand{\cello}{\cellcolor{orange!20}}

\newcommand{\john}[1]{}
\newcommand{\michael}[1]{}
\newcommand{\ryan}[1]{}
\newcommand{\stephen}[1]{}
\newcommand{\lucy}[1]{}
\newcommand{\kathryn}[1]{}

\newcommand{\figblock}[1]{\emph{#1}}

\makeatletter
\DeclareRobustCommand\onedot{\futurelet\@let@token\@onedot}
\def\@onedot{\ifx\@let@token.\else.\null\fi\xspace}

\def\etal{\emph{et al}\onedot}
\makeatother

%% file: main_sections/01_intro.tex
\section{Introduction}
\label{sec:introduction}

In recent years, approaches to view synthesis have achieved impressive photorealistic results~\cite{mildenhall2020nerf,mueller2022instant,kerbl_2023}. These algorithms generally require two steps to generate novel views of a scene. First, they perform an optimization procedure to reconstruct a 3D physical representation. Second, the 3D representation is rendered from a novel view. While some approaches are capable of performing the rendering step in real-time, the reconstruction step remains stubbornly slow.

We present a neural algorithm that achieves real-time rates for reconstruction and rendering \emph{combined}. As depicted in Fig.~\ref{fig:teaser}, our algorithm takes as input an array of wide baseline high-resolution images or video streams and produces high-quality novel view renderings. 
In our experiments (see Sec.~\ref{sec:results}), we use 8 input images per frame that are as far as 30cm apart, and we demonstrate that our model can produce novel views at 30fps at 1080p (1920$\times$1080) resolution on an NVIDIA A100 GPU. Through thorough qualitative and quantitative analysis, we demonstrate state-of-the-art quality at real-time rates. More impressively, our quality approaches some of the top offline methods, and in some cases surpasses them. Through extensive ablations we validate our design choices and show that our network is highly tuneable, and can achieve even higher quality if slower (e.g. 10fps) rendering is acceptable.

Our approach combines several key concepts that are described below and highlighted in Fig.~\ref{fig:teaser}.

\paragraph{Layered depth map (LDM) scene representation.}
As depicted in the top-right of Fig.~\ref{fig:teaser}, we make use of a layered depth map (LDM) 3D scene representation. The output LDM uses a small number (6 for our fastest model) of layers, each with an associated depth map, density map, and blend weights. The depth map geometry conforms to objects in the scene, the density map models occlusions and anti-aliased edges, and the blend weights blend over the input image pixels to produce high-resolution output images. Note that the LDM is closely related to a layered mesh (LM) from previous works~\cite{broxton2020immersive, solovev2023, khakhulin2022stereo}, but, as described in Sec.~\ref{sec:method}, we never need to instantiate a mesh from our depth maps. 

Moreover, our method reconstructs, renders, and discards the LDM for every frame. Hence we can optimize the LDM to each specific novel view in a video sequence, aligning it with the view; generating depth, density, and blend weights that are optimized for that view; and rendering with a simple pixel-aligned over operation. As demonstrated in our results (Sec.~\ref{sec:results}), this especially helps for scenes with reflective and refractive materials which our LDM representation doesn't model explicitly.

\paragraph{Multi-scale learned render-and-refine network core.}
In order to create an efficient network to solve the LDM for each frame, we use a multi-scale learned render-and-refine network structure as highlighted in the bottom of Fig.~\ref{fig:teaser}. The learned render-and-refine approach was first introduced by~\cite{flynn2019deepview} and is similar to an unrolled gradient descent but with dramatically faster convergence properties (e.g. 5 iterations instead of thousands). As described in Sec.~\ref{sec:method}, in each \figblock{Update \& Fuse} step in Fig.~\ref{fig:teaser} the network renders the current LDM estimate to each of the input views, and uses the result to refine the LDM. 
Still, previous implementations of learned render-and-refine~\cite{flynn2019deepview, broxton2020immersive, solovev2023} are far too slow for real-time.

Our work is the first to make this learned render-and-refine approach real-time, which we achieve by embedding the \figblock{Update \& Fuse} steps into a UNet~\cite{ronneberger2015u} structure. As shown in Fig.~\ref{fig:teaser}, the encoded image features are first projected into a multi-scale pyramid. Then the first \figblock{Update \& Fuse} step starts at the lowest, aggressively down-scaled resolution. Each successive step improves the LDM solution, while some increase the spatial resolution (the second, fourth, and fifth in Fig.~\ref{fig:teaser}), and some decrease the number of LDM layers (the last two in Fig.~\ref{fig:teaser}). The number of layers progresses from high to low because the depth dimension for most scenes is best represented by a small number of impulses (surfaces). The denser depth sampling at early iterations locates those surfaces, and the fewer layers at later iterations more closely follow them. As shown in our ablations (Sec.~\ref{sec:results_ablations}), reducing the layers slightly reduces quality, but the benefit in significantly reduced runtime makes it well worth it. This network architecture efficiently balances the compute between the spatial and depth dimensions.

To further optimize the network, the final \figblock{Update \& Fuse} generates an LDM at a reduced resolution (scaling factor $s$ in Fig.~\ref{fig:teaser}, which is $\sim$2$\times$--4$\times$ in our experiments), and we upscale the unactivated LDM attributes (depth, density, and blend weights) with a simple bilinear upsample followed by an activation (\figblock{Upsample \& Activate} in Fig.~\ref{fig:teaser}).
This approach has been demonstrated to be effective for piecewise smooth functions, like our LDM attributes, as it cleanly interpolates the smooth regions while maintaining sharp edges~\cite{ReluField_sigg_22}.

\paragraph{Transformer-based input view fusion.} A central problem in view synthesis is how to aggregate information from multiple views in an efficient and order-independent manner. As  described in Sec.~\ref{sec:method_attention}, our method
solves this by incorporating a Transformer~\cite{vaswani2017attention}-based network component within each \figblock{Update \& Fuse} step.
Similar approaches have proven effective in previous work~\cite{Yang_2019,reizenstein2021common} but can be expensive. We introduce a novel, optimized variant of cross-attention, \textit{One-to-many attention}, that dramatically lowers computational requirements.

We additionally show how to replace a Transformer's traditional positional encoding with a directional encoding based on the pose of the input images.

\paragraph{Results.} When taken together, these, along with many other smaller design choices fully described in Sec.~\ref{sec:method} and justified with extensive ablations in Sec.~\ref{sec:results_ablations}, produce a novel algorithm that produces high-quality synthesized images at real-time rates as demonstrated by our results in Sec~\ref{sec:results}. 

%% file: main_sections/02_background.tex
\section{Related Work}
\label{sec:related}
Our approach aims to solve the problem of \emph{novel view synthesis}, where, given a set of input images from different views of a scene, we generate a novel view that reproduces complex 3D parallax effects and occlusions. Most classical solutions to this problem can be classified as \emph{image-based rendering (IBR)}~\cite{chen1993,mcmillan1995}. More recent solutions that use neural networks are considered \emph{neural rendering} algorithms~\cite{Tewari2022NeuRendSTAR}. Ours is a neural rendering approach that uses IBR in order to efficiently produce high-quality, high-resolution results. Here we focus on the more recent neural rendering efforts; the classical methods are well covered by surveys~\cite{wu_2017,zhang_2004}. 

The main aspects of our approach that most clearly distinguish it from others are its (1) LDM \emph{physical representation} (Sec~\ref{sec:prior_rep}) and (2) novel \emph{generalizable} and efficient neural algorithm (Sec.~\ref{sec:prior_alg}).

\subsection{Physical Representations for View Synthesis} \label{sec:prior_rep}
The quality of view synthesis algorithms is highly dependent on the accuracy of their physical representation~\cite{chai2000plenoptic,lin2004geometric}. Recent view synthesis approaches use a wide variety of representations, including implicit surfaces~\cite{Kellnhofer:2021:nlr,oechsle2021unisurf,wang2021neus,Niemeyer_2020_CVPR,Sitzmann_2019}, point clouds~\cite{aliev2020neural, Lassner_2021_CVPR, ruckert2022adop, Wiles_2020_CVPR,yifan2019differentiable}, voxels~\cite{Sitzmann_2019_CVPR,liu2020neural,yu2021plenoctrees,Hedman_2021_ICCV,fridovich2022plenoxels}, 3D Gaussians~\cite{kerbl_2023}, triangle surface meshes~\cite{Burov_2021_ICCV,Hu_2021_ICCV,thies_2019},  multi-plane images (MPIs)~\cite{flynn2019deepview, mildenhall2019llff,Wizadwongsa_2021_CVPR,tucker2020single}, multi-sphere images (MSIs)~\cite{broxton2020immersive, attal_2020}), layered meshes (LMs)~\cite{broxton2020immersive, khakhulin2022stereo, solovev2023} and volumetric ray marches of neural fields~\cite{mildenhall2020nerf,barron2021mipnerf,mueller2022instant}. 

The highest quality algorithms tend to use fully volumetric representations, many with a volumetric ray march~\cite{mildenhall2020nerf}, whereas the surface based representations, like implicit surfaces and surface meshes, are tailored for efficient rendering. Volumetric representations better capture complex geometry with many interocclusions, like leaves.
Unfortunately, these methods can be slow, especially for scene reconstruction. There are many recent approaches to speed up the ray march
\cite{mueller2022instant, adaptiveshells2023,lin2022efficient,barron2023zipnerf,wan2023learning,Reiser_2021_ICCV,yariv2021volume,Garbin_2021_ICCV,tensorf_chen_2022,im4d_lin_2023,Attal_2023_CVPR,Fridovich-Keil_2023_CVPR}, but we're only aware of one (eNeRF~\cite{lin2022efficient}, discussed in more detail below) that can achieve interactive rates for combined reconstruction and rendering while also producing competitive quality for some scenes.

In our work, we use an LDM representation which is similar to an LM. Previous work that use LMs also internally solve an LDM and then map to an LM in order to reproject to other views, but as described in Sec.~\ref{sec:method} we render the LDM directly without a mesh. LDMs (and LMs) aim to combine the quality of fully volumetric representations and the efficiency of surfaces. The layers can be considered steps of a volumetric ray march, while the depth map within each layer follows the smooth surfaces that make up the largest portion of real-world scenes. While Gaussian splats~\cite{kerbl_2023} have emerged as a promising new representation enabling fast rendering, their computation, like that of NeRF, requires a slow, offline, per-scene process. The 2D surfaces of the LDM allow us to use efficient 2D network components to solve the scene. One of the historical downsides of LDMs and LMs is that they can't be viewed from all directions, but, as demonstrated in Sec.~\ref{sec:results}, this is less of a limitation for our method because we regenerate the LDM for every novel view.

\subsection{Generalizable Neural Algorithms for View Synthesis} \label{sec:prior_alg}
While the physical representation determines the asymptotic quality limit for novel view synthesis, the algorithm determines both how closely it approaches that limit and the overall speed. As described in Sec.~\ref{sec:method}, we use a novel architecture that is designed for efficiency, specifically during scene reconstruction, and generalizability, to reconstruct a broad range of scenes using a small number of input images.

Many recent approaches use a non-generalizable offline optimization, usually gradient descent, in order to fit the scene to their physical representation~\cite{mildenhall2020nerf,barron2021mipnerf,kerbl_2023}. These methods produce high-quality results, but they must be separately trained for every scene, which, while an active area of research, is still a slow, iterative process. Moreover, they require a large and dense set of input images in order to achieve high quality.

Our approach is \emph{generalizable}, which means both that it doesn't have to be trained per scene and also that we can produce high quality even with a small number of wide-baseline images. There are several other recent generalizable neural rendering methods~\cite{suhail2022generalizable,yu2020pixelnerf,wang2021ibrnet,mvsnerf,Cao2022FWD,Jain_2021_ICCV}. We directly compare to three state-of-the-art methods in our results (Sec.~\ref{sec:results}), and we describe them in more detail later in this section.

A central problem for generalizable view synthesis networks is how to aggregate information from multiple views. 
Earlier methods~\cite{flynn:2016:deepstereo,mildenhall2019llff} used simple concatenation of across-view features, but they are not invariant to the input camera ordering, and they must be trained for a specific number of input images.
Recurrent networks~\cite{choy20163dr2n2,kar2017learning} allow for variable numbers of input images but are also not order-invariant. Statistical measures such as mean~\cite{yu2020pixelnerf} and variance~\cite{wang2021ibrnet} are simple and order-invariant, but have limited representational power. 
\cite{flynn2019deepview}, on the other hand, uses repeated max-pooling, which has been shown \cite{qi2017pointnet} to be a powerful set aggregator.  \ryan{Does this one have a downside? If not, why don't we use it instead of our Transformer-based approach?}

With the increasing popularity of Transformers \cite{vaswani2017attention}, attention has emerged as a natural alternative \cite{lee2018set}. 
Earlier work, the \emph{AggTransformer}~\cite{Yang_2019} uses a simpler form of attention where each feature is projected to an attention weight, and these weights are aggregated with a \emph{softmax}, but it does not repeat this operation, which limits its power.
In contrast, the \emph{NerFormer} \cite{reizenstein2021common} repeatedly applies self-attention across the $M$ views, alternating with attention along the ray. This method retains a Transformer's full expressive power, but is expensive due to self-attention's well-known $O(M^2)$ complexity. Our proposed method, \emph{One-to-many attention}, aims to combine the efficiency of the AggTransformer while largely retaining its expressiveness.

In our results (Sec.~\ref{sec:results}), we directly compare to the following three recent generalizable neural view synthesis algorithms. Out of the most recent state-of-the-art methods, these three had the most similarities to ours.

GPNR~\cite{suhail2022generalizable} also aims to produce state-of-the art quality with a generalizable neural network. 
GPNR extracts patches from neighboring images along the novel view rays' epipolar lines and feeds them into a Transformer architecture to solve the color for each ray. We also use a Transformer-inspired component in our network. However, as described in Sec.~\ref{sec:method_attention}, we use a novel optimized form of One-to-many attention, and as shown in Sec.~\ref{sec:results}, our quality is competitive with and often surpasses GPNR, even though our method is over $700\times$ faster.

As of the time of this paper, eNeRF~\cite{lin2022efficient} is the fastest generalizable neural rendering approach that also achieves competitive quality for some scenes. They speed up ray marching by sampling only around a single surface, represented by a depth map with varying thickness. 
eNeRF runs at interactive rates at lower resolutions and achieves competitive quality for simpler scenes with limited depth complexity. As shown in our results, we achieve higher quality for more complex scenes while being at least 6$\times$ faster.

We also compare to SIMPLI~\cite{solovev2023} because their algorithm has commonalities with ours. They also use a render-and-refine structure and an LDM geometry. %
However, as shown in Sec.~\ref{sec:method}, our network differs in critical ways, most significantly our multi-scale LDM architecture and our Transformer-inspired One-to-many attention module for view fusion. Our results show that our quality surpasses SIMPLI while running over $100\times$ faster.

\subsection{Single-view Methods}
There is a growing class of view synthesis approaches that can render 3D scenes from very restricted views, a monocular video stream, or even a single image. These include approaches that use generative algorithms with a learned prior~\cite{trevithick2023, Chan_2022_CVPR,Xu_2022_SinNeRF,tseng2023consistent,liu2023zero,Du_2023_CVPR}; approaches that use a more explicit prior, usually a human head~\cite{Wang_2021_CVPR,meshry2021learned,Ma_2021_CVPR,chu2020expressive,lombardi2021mixture,peng2021neural} or full body~\cite{peng2021animatable,remelli2022drivable}; and approaches that use multiple views over time~\cite{Li_2021_CVPR,Tretschk_2021_ICCV,park2021hypernerf,park2021nerfies,pumarola2021d,Xian_2021_CVPR}. This is an exciting space that can enable simpler image capture hardware and processes. 

However, the multi-view approaches discussed in Secs.~\ref{sec:prior_rep} and~\ref{sec:prior_alg} currently provide significantly higher quality and generalize to a broader range of scenes.
Thus we consider these single-view methods a separate class of algorithm and do not compare directly to them in this paper.

%% file: main_sections/03_method_v2.tex
\section{Method}
\label{sec:method}
In this section, we describe the key components of the Quark algorithm shown in Fig.~\ref{fig:teaser}. In order to achieve high quality at real-time rates, we structure our network to perform a multi-scale iterative render-and-refine process that allows most processing to be performed at low resolution. First, the input pixel colors and ray directions in the \figblock{Input Views} are encoded (\figblock{Encode Input Images}) into a feature pyramid~\cite{Lin_2017_CVPR}. Next, these multi-scale features are fed into \figblock{Iterative Updates}, which are a series of \figblock{Update \& Fuse} steps that refine the LDM. These steps save compute by balancing the increase in the LDM's resolution with a decrease in the number of layers. To guide the refinement within each \figblock{Update \& Fuse} step (shown in expanded detail in Fig.~\ref{fig:update-step}), the LDM is rendered to the input views (\figblock{Render to Input Views}). We aggregate features from the different views using an efficient One-to-many attention mechanism. 

When the process of \figblock{Iterative Updates} described above concludes, a final LDM is produced with dimensions that are a scale factor $s$ smaller than the desired output resolution. The final LDM can then be rendered to a full resolution novel view via the following lightweight process. First, \figblock{Upsample \& Activate} bilinearly interpolates the LDM attributes (depth, density, and blend weights) to the output resolution and activates them. Next, the original input views are projected onto the depth layers, these are combined into a single image per layer with blend weights, and then layers are over-composited back to front to produce the rendered novel view. If needed, a depth map can optionally be produced by over-compositing the decoded depth instead of the blended input views.

The following subsections describe these stages in more detail. We begin with the LDM output representation and rendering equation in Sec.~\ref{sec:method-ldm}, followed by an end-to-end description of the algorithm that generates the LDM in Section~\ref{sec:method-solving}. We expand upon our Transformer-based fusion mechanism in Sec.~\ref{sec:method_attention}, and we detail our training procedure in Sec.~\ref{sec:method_training}.

\subsection{The Layered Depth Map (LDM)}\label{sec:method-ldm}
In this section we will describe the LDM produced at the output of the Quark network and show how it can be used to render the final RGB image. The LDM  consists of a series of $L$ layers with spatial dimensions $[H, W]$ that are situated within the frustum of the novel viewpoint being rendered, which we henceforth refer to as the \emph{target} viewpoint. LDM layers have three associated attributes: depths $\mathbf{d}$, densities $\mathbf{\sigma}$, and blend weights ${\blendweights}$ (see Figure~\ref{fig:teaser}). \ryan{I don't think we need the next 2 sentences. They feel redundant.} The depth and density contain the $[L, H, W, 1]$ depth and transparency (i.e. alpha) values respectively. The blend weights ${\blendweights}$ contain $[L, H, W, M]$ coefficients for blending $\numimages$ input images on each layer.

To render the target image from an LDM, we first back-project the input images onto the depth layers. We define the operator $\mathscr{P}^T_\mathbf{\theta}(\,\mathbf{I},\;\mathbf{d})$ for this purpose (the transpose here denoting that this is the adjoint of the normal forward projection operator $\mathscr{P}$ that will be introduced shortly).  Here $\mathbf{I}$ is an $[M, H, W, C]$ tensor of input images, and $\theta$ are the camera parameters. The back-projected input images are blended using the per-image blend weights $\beta_m$ to create per-layer RGB. This RGB, along with the density $\sigma$, is then over-composited to produce the final image. Let $\mathscr{O}:(\mathbf{c},\mathbf{\sigma})\mapsto\textbf{I}$ be the standard over composite operator, which renders an image by alpha compositing the appearance $\mathbf{c}$ at each layer from back to front. The resulting render is:
\begin{equation}
\mathrm{\mathbf{c}_\mathrm{target}}
= \mathscr{O}\left( \sum_{m=1}^{M} \beta_m \cdot \mathscr{P}^T_\mathbf{\theta}(\,\mathbf{I},\;\mathbf{d}), \mathbf{\sigma} \right).
\end{equation}
While training leverages standard differentiable rendering components, during inference we use a CUDA-optimized renderer, enabling 1080p resolution at approximately 1.3 ms per frame.
\begin{figure}
    \centering
    \includegraphics[width=0.6\columnwidth]{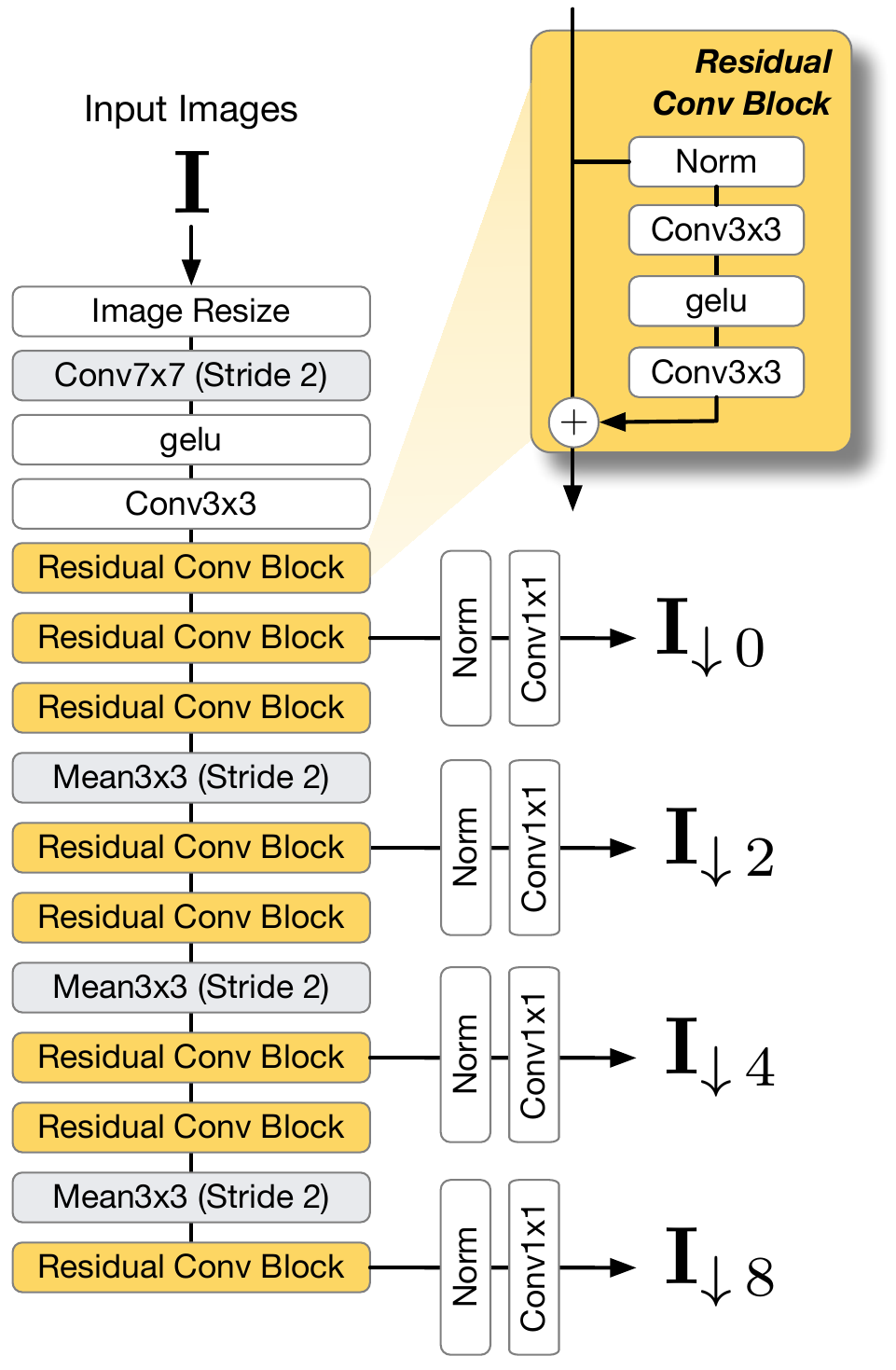}
    \caption{\textbf{Encode Input Images} from Fig.~\ref{fig:teaser}. Quark encodes and downsamples input images using a series of residual networks and strided mean-pooling layers.}
    \label{fig:input-encoder}
\end{figure}

\subsection{The Quark Algorithm}\label{sec:method-solving}
\begin{figure}
    \centering
    \includegraphics[width=\columnwidth]{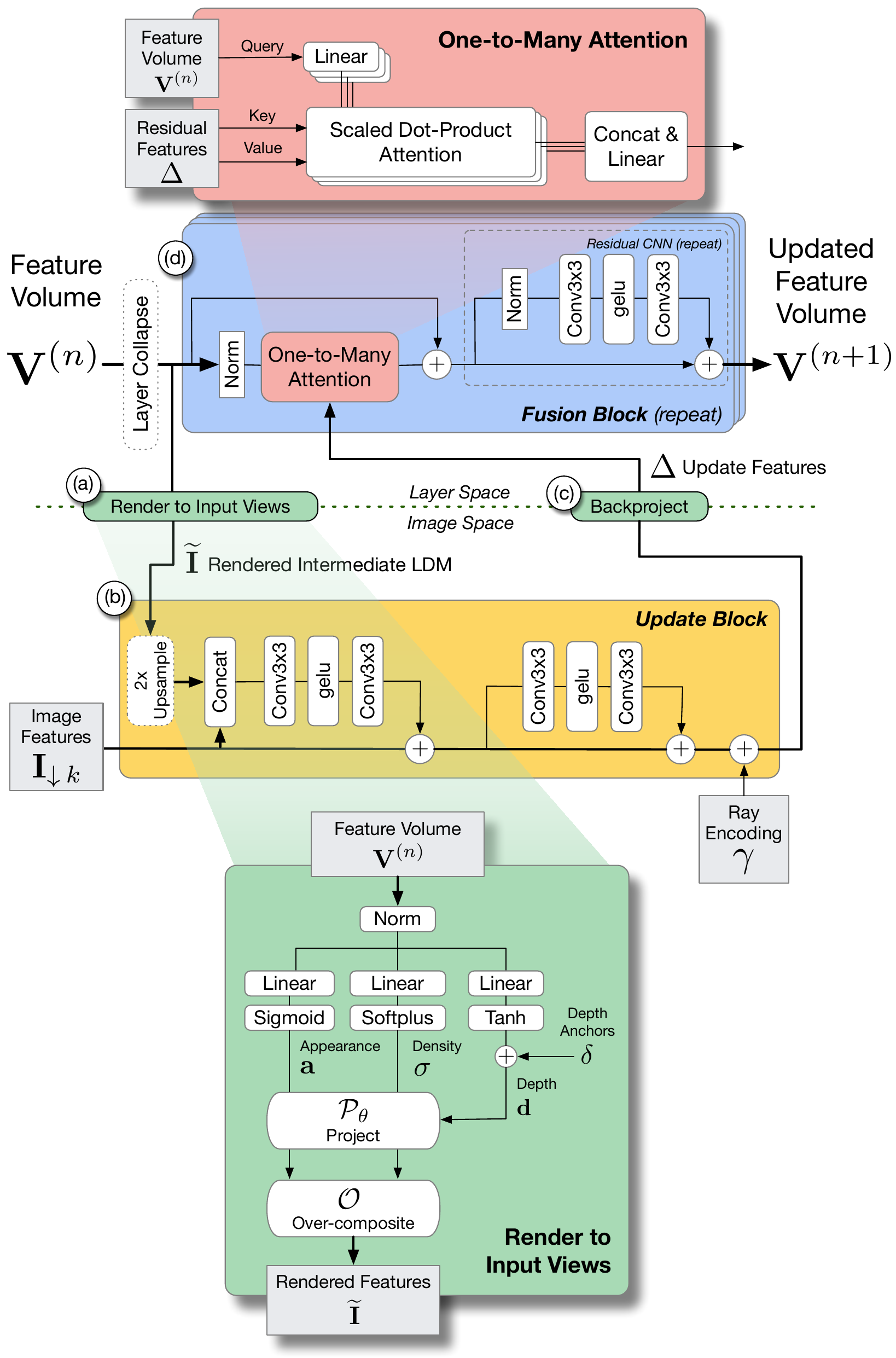}
    \caption{\textbf{Update \& Fuse Step} from Fig.~\ref{fig:teaser}. During each iteration the \figblock{Update \& Fuse} step uses a render-and-refine approach to generate a refined feature volume. \textbf{(a)} First, the feature volume is decoded into an LDM and rendered $M$ times into each of the input viewpoints (see bottom inset). \textbf{(b)} Next, the rendered features are combined with input features $\mathbf{I}_k$ and encoded ray dirs $\gamma_k$ via a residual \figblock{Feed-forward CNN} to generate update features from each view. During iterations where the feature volume is upscaled, the rendered intermediate LDM is upsampled by a factor of two in the spatial dimension and combined with image features at the next level of detail. \textbf{(c)} Updated features are \figblock{Backproject}ed into the feature volumes using the same depths $d$ decoded in (a). \textbf{(d)} Finally, updates from all views are combined into a single set of update features $\Delta$ and fed into the \figblock{Fusion block}, which uses across-view attention (top inset) to reason about visibility and update the feature volume. Note that \figblock{Fusion Block} is repeated a variable number of times during each iteration, as is the residual CNN within it. Layer collapse, which reduces the number of layers by a factor of 2 via a residual CNN, is also applied during the final two iterations. See Tabs.~\ref{table:supp_quark} and~\ref{table:supp_quarkplus} for these and other per-iteration implementation details.}
    
    \label{fig:update-step}
\end{figure}

In this subsection, we describe the algorithm's three major sub-components, with a particular focus on the iterative multi-scale render-and-refine approach. To aid in understanding, we encourage the reader to refer to Fig.~\ref{fig:teaser}, which shows the overall network structure, and Fig.~\ref{fig:update-step}, which shows the \figblock{Update \& Fuse} step in detail.

As motivation for our approach, we note that solving the LDM directly at the final output resolution $[H, W]$ would be computationally expensive. Instead, our method solves for the final high resolution LDM by first aggressively downsampling and encoding $\numimages$ input images, and then iteratively refining the LDM over $N$ render-and-refine update steps that progressively increase spatial resolution while decreasing the number of layers (see bottom row of Figure~\ref{fig:teaser}). This multi-scale approach is fundamental to the speed of the Quark network; in early iterations, computation with more layers is performed at very low spatial resolution, and in later iterations, the cost of high spatial resolution is offset by layer reduction.

\subsubsection{Encode Input Images.} \label{sec:method-encode-input} The $\numimages$ input images are first converted to low resolution feature map pyramids (each with $K$ levels) using a standard convolutional encoder (see Fig.~\ref{fig:input-encoder}).  This encoding step allows the model to operate at reduced resolutions, rather than on the input images directly. At each update step $n$ in the solver, we choose the appropriate resolution feature map from these $K$ feature maps.

For each image feature map $\mathbf{I}_k$, we additionally encode the direction of image rays relative to the target camera using a ray directional encoding $\gamma_{k}$. The encoded ray directions allow the network to bias towards input views that are closer to the target view, leading to improved results for reflections and non-Lambertian surfaces (as can be seen in Sec.~\ref{sec:results} and Sec.~\ref{sec:results_ablations}).  Several methods to encode the ray direction have been proposed. \cite{mildenhall2020nerf} uses a sinusoidal positional encoding \cite{vaswani2017attention,tancik2020fourfeat}, while ~\cite{verbin2022refnerf} use spherical harmonics.
While these methods work well for general scenes, the geometry of the LDM allows for a simpler ray directional encoding. Specifically, we encode a ray by first computing the difference vector between the ray's intersections with the near and far planes of the LDM frustum in projective space. After applying a \textit{tanh} non-linearity to this 2D difference vector, we encode it with a sinusoidal positional encoding. The resulting ray encoding method has several desirable properties. Firstly, it evaluates to zero when an input ray exactly aligns with an LDM (and target view) ray. Secondly, the precision is concentrated on ray directions of interest, namely those intersecting the near and far planes within or close to the LDM frustum bounds. Finally, the \textit{tanh} non-linearity ensures that rays falling outside the frustum are still represented, albeit with less precision.

For efficiency, the ray directional encoding is computed once at low resolution and bilinearly upsampled to match the dimensions within the feature map pyramid. We use 8 octaves of sinusoidal encoding and project the result using a linear layer to $C$ channels. A separate projection matrix is applied for each required resolution.

\subsubsection{Iterative Updates} The network increases from coarse to fine resolution over a series of five \figblock{Update \& Fuse} steps. During the solve, the LDM is processed in an encoded form $V^{(n)}$ that we refer to as the \emph{feature volume} (see Fig.~\ref{fig:teaser}). Note that both the number of layers $L$ and spatial resolution of the feature volume change across the update steps; we increase the resolution while simultaneously decreasing the number of output layers, reminiscent of a UNet~\cite{ronneberger2015u}. The number of channels is held constant across iterations.

During each iteration, the solver calls an \figblock{Update \& Fuse} network block, shown in expanded detail in Fig.~\ref{fig:update-step}. Each \figblock{Update \& Fuse}  block decodes $V^{(n)}$ to produce an \emph{intermediate LDM} by extracting depth, density, and appearance features. Next, we bilinearly splat the per-layer density and appearance into each input view. Splatting avoids expensive ray tracing or rasterization. Finally, we over-composite these projected layers to create the rendered image at each input view. This entire process appears as \figblock{Render to Input Views} in Fig.~\ref{fig:update-step}.

The rendered LDM $\widetilde{\mathbf{I}}$ is then combined in the \figblock{Update Block} with the corresponding input image features $\mathbf{I}_k$ and encoded ray directions $\rho_k$, selected from the $k$-level image feature pyramid. Note that the \figblock{Update Block} may include an upsampling of $\widetilde{\textbf{I}}$ if there is a transition between resolutions for this step. The \figblock{Update Block} then produces \emph{update features} using a small CNN, and back-projects these into the feature volume. This per-view array of update features $\Delta$ encodes both extracted image features (hence it bears some similarity to a plane sweep volume) as well as visibility information derived from comparing $\widetilde{\mathbf{I}}$ to $\mathbf{I_k}$. The per-view update features are aggregated by a series of \figblock{Fusion Blocks}. Each fusion block applies One-to-many attention to fuse the back-projected image features $\Delta$ followed by one or more residual CNNs to update the encoded LDM $V$. 

The remainder of this section describes the \figblock{Update \& Fuse} network in more detail. The attention mechanism in the \figblock{Fusion Block} is described in Section~\ref{sec:method_attention}.

\paragraph{Decoding and rendering the intermediate LDM:}
To render the intermediate LDM (in \figblock{Render to Input Views} in Fig.~\ref{fig:update-step}),
we first decode the feature volume $\mathbf{V}^{(n)}$ to instantiate the intermediate LDM. We decode density and depth in a similar way as for the final LDM, which we describe below in Sec.~\ref{sec:upsampling-and-activation}. 
However, for efficiency, we activate before projecting to the views. Additionally, rather than using blend weights we instantiate the intermediate LDM's appearance directly via $\mathbf{c}=\mathrm{sigmoid}(\mathbf{V}\, W_\mathbf{a})$.
When rendering, we project the intermediate LDM layers through the depth map $\mathbf{d}$ into each of the input views using the projection operator $\mathscr{P}_\mathbf{\theta}:(\textbf{V},\mathbf{d})\mapsto \{\textbf{I},\dots \}$, and over-composite the result to yield: 
\begin{equation} \label{eq:overcomposite-result}
{\widetilde{\textbf{I}}} =
\mathscr{O}\big(
  \mathscr{P}_{\mathbf{\theta}}(\mathbf{a}, \mathbf{d}), \;
  \mathscr{P}_{\mathbf{\theta}}(\mathbf{\sigma}, \mathbf{d})\big)
\end{equation}  
 for that input view. %

\paragraph{Layer Collapse:} Prior to each of the final two \figblock{Update \& Fuse} steps, we reduce the number of layers by a factor of two using a simple residual MLP, where the straight through path is the mean of adjacent layers, and the residual path is their concatenation. This reduction in layers, for a small reduction in quality (see ablations in Sec.~\ref{sec:results_ablations}), dramatically reduces computation at these higher resolution iterations. 

\paragraph{Initialization:} Initialization of the LDM (\figblock{Learned Initialization} in Fig.~\ref{fig:teaser}) uses a special case of the \figblock{Update \& Fuse} step. Since there is no existing LDM at the first iteration, we instead start with a single learned $C$-channel feature broadcasted to initial spatial dimensions $H^{(0)}, W^{(0)}$.  Additionally, we assume the depth layers are flat (set to predefined depth anchor values, explained below in Sec.~\ref{sec:upsampling-and-activation}) and omit the rendered image $\widetilde{\mathbf{I}}$, leading to a simplified \figblock{Update Block} that focuses on combining image features and ray directions.

\subsubsection{Upsampling \& Activation.}  
\label{sec:upsampling-and-activation}

After the \figblock{Update \& Fuse} steps, the feature volume contains the final scene representation, albeit in an encoded form and still at relatively low resolution. The final LDM is decoded from $V^{(N)}$ using linear projections followed by non-linear activation functions (\figblock{Upsample \& Activate} in Fig.~\ref{fig:teaser}). Specifically, density is decoded using $\mathbf{\sigma}=\mathrm{sigmoid}(\mathbf{V}\, W_\mathbf{\sigma})$, where $W_\mathbf\sigma$ are linear weights learned by the network. 
When we decode depths $\mathbf{d}$, we constrain the resulting LDM layers to equally spaced disparity bands within the near and far plane ($\eta$ and $f$ respectively) of the target view frustum.
To do this, we compute depth relative to \emph{depth anchors} $\delta$ which are just the center point of each band. Thus for each layer, $\ell = 1, \dots, L$:

\begin{equation}
\delta = \frac{\ell-0.5}{L}
\end{equation}

and the final depth $\mathbf{d}$ is: 
\begin{equation} \label{eq:depth_activation}
\mathbf{d_{\ell}} = \left[\left(\delta  + \frac{0.5}{L} * \mathrm{tanh}(\volume\, W_\mathbf{d})\right) \cdot \left( \frac{1}{\eta} - \frac{1}{f} \right) + \frac{1}{f} \right]^{-1}
\end{equation}

We compute the blend weights $\mathbf{\beta}$ using the same One-to-many attention mechanism used within the \figblock{Fusion Block}s. However, we delay applying the \textit{softmax} until after upsampling to the target resolution. This step is incorporated into the final fusion block and uses that block's image update features $\residual$.

All activations are applied after bilinear upsampling to the final target resolution. For real scenes, depth and density are typically piece-wise smooth, and post-sampling activation \cite{ReluField_sigg_22,SunSC22} can produce crisp occlusion boundaries, while blending over the full resolution input images produces high resolution RGB output.

\subsection{One-to-Many Attention in the Fusion Block}\label{sec:method_attention}

In this subsection, we describe the across-view attention mechanism that fuses the update features $\residual$ to produce an updated feature volume $\mathbf{V}^{(n+1)}$ in the \figblock{Fusion Block} in Fig.~\ref{fig:update-step}.

We use a novel \figblock{One-to-many attention} block that aggregates per-view update features $\residual$  by repeatedly cross-attending from a single aggregated feature to our $M$ per-view update features. Intuitively, this attention mechanism (softly) selects the most relevant information from the per-view update features $\residual$, implicitly incorporating both image matching cues and occlusion. Echoing a Transformer's  \cite{vaswani2017attention} positional encoding, the ray directional encoding included within the update feature allows the network to bias toward rays closer to the target view. Our method inherits cross-attention's $O(M)$ complexity, but by exploiting specific redundancies in the Transformer's formulation, the constant factor can be greatly reduced for our typical input sizes, leading to an algorithm that is closer to $O(1)$ (see Appendix~\ref{sec:app_attention} for a complexity analysis).

\paragraph{Preliminaries.} Given a set of queries $Q$, keys $K$ and values $\text{\emph{Val}}$,  the standard attention operation computes the output according to 
\begin{equation}
\text{Attention}(Q, K, \text{\emph{Val}}) := \text{softmax}\left(\frac{QK^T}{\sqrt{C}}\right)\text{ \emph{Val}},
\end{equation}
where $C$ is the number of channels. 

The full Transformer uses multi-headed attention where the query, keys, and values are each projected to $h$ heads, each with $C/h$ channels, before performing $h$ attention queries. The resulting values are then concatenated and projected to the final output. This operation is performed multiple times, interleaved with MLP blocks. 

We adapt the Transformer to aggregate across view features. Specifically, the current $\volume$ cross-attends over the $\numimages$ update features ($\residual$). In standard cross-attention the queries and keys for the different attention heads are derived by linear projection from both of these values. That is,
\begin{equation}
\text{MultiHead}(\volume, \residual) = \text{concat}(\text{head}_{1}, \dots, \text{head}_{h})W^O
\end{equation}
with
\begin{equation}
\text{head}_i = \text{Attention}\left(
\volume W_i^q, \residual W_i^k, \residual W_i^{val}
\right)
\end{equation}
and $W^O$ the output projection.

\paragraph{Optimized Attention.}  The standard multi-headed attention operation requires $h \times M$ matrix multiplies, each of size $C \times C/h$, to multiply the $\residual$s with $W_i^k$ and $W_i^{val}$. However, as noted in  \cite{turner2024introduction} (side-note 13), and described in detail in the appendix, there is a redundancy in the standard attention formulation; under certain conditions it is mathematically equivalent, but much more efficient, to omit the matrix multiplies on $\residual$ and instead fold them into $W_i^q$ and $W^O$, leading to the following formulation:
\begin{equation}
\text{head}_i = \text{Attention}\left(
\volume W_i^{q}, \residual, \residual
\right)
\end{equation}

This removes all of the matrix multiplies to produce the $M$ keys and values, leaving only the dot products and sums. As shown in the analysis in Appendix~\ref{sec:app_attention}, this is considerably more efficient for the number of inputs and heads used in our attention blocks. Additionally, we gain a further efficiency by reducing expensive layer-space computation. Standard cross attention would require repeated (for multiple attention rounds) matrix multiplies over the $M$ per-view update features on each of the $L \times H \times W$ LDM elements, requiring $O(LM)$ matrix multiplies per element. In contrast, our optimized One-to-many attention method eliminates the need for \emph{any} matrix multiplications on the update features within layer-space (i.e $O(LM)$ matrix multiplies); instead, the majority of per-view computation occurs in image space and is lifted to 3D.

\paragraph{Fusion Block.} 
A pre-LayerNorm Transformer \cite{xiong2020layer} consists of residual self attention blocks followed by residual MLP blocks. Following that formulation, we define the full One-to-many attention block, including normalization \cite{zhang2019root}, $\Norm$, as
\begin{equation}
\mathbf{V'} = \volume + \text{MultiHead}(\Norm(\volume), \residual).
\end{equation}

Mirroring a Transformer's MLP block, the One-to-many attention module is interleaved with $2D$ 3$\times$3 residual convolutional blocks:
\begin{equation}
\mathbf{V'} = \volume + \text{Conv}(\text{gelu}(\text{Conv}(\Norm(\volume))).
\end{equation}

The full \figblock{Fusion Block}, consisting of One-to-many attention and convolution blocks, is repeated several times within each \figblock{Update \& Fusion} block and learns to both aggregate across views and to constrain the LDM to the manifold of real scenes.

\subsection{Training}
\label{sec:method_training}

Similar to other generalizable view synthesis methods ~\cite{flynn2019deepview,suhail2022generalizable,lin2022efficient,solovev2023}, we train on a collection of calibrated multi-view images across a variety of scenes.
Specifically, the network is trained with 8 input views rendered to a held out target image. The 8 input views are randomly selected from the sixteen views closest to the target. We train our model using a combined $10* L1 + LPIPS$~\cite{Zhang_2018_CVPR} loss function, with batch size 16 across a single 16 A100 machine. 
We train our models for 250K iterations on half resolution input images, with a similarly half resolution encoded LDM, before increasing to full resolution for 350K steps, for a total of 600K steps.
To reduce RAM consumption and training time we randomly crop the target and rendered image to size 256$\times$256 during initial low resolution training and 512$\times$512 pixels during high res training. The learning rate is warmed up to $1.5\times10^{-4}$ over 20k iterations, held constant for 520K iterations and then cosine-decayed to zero over the final 680K iterations. We use residual scaling \cite{radford2019language} throughout our model to improve stability.

\input{figures/table_comparison.tex}

We trained our models on a weighted combination of the \textit{Spaces} \cite{flynn2019deepview}, \textit{RFF}~\cite{mildenhall2020nerf}, \textit{Nex-Shiny} ~\cite{Wizadwongsa_2021_CVPR} and {SWORD}~\cite{solovev2023} datasets. All datasets had equal weighting apart from the much smaller \textit{Nex-Shiny} dataset which was weighted at 0.25. These datasets have different image resolutions, so for each example we adjust the aspect ratio of the internal LDM dimensions within our model accordingly, retaining the same total area per layer and thus the same computational requirements as our 1080p model. Similarly, we chose a different near and far plane adaptively for each example, based on the distribution of depths from the available reconstructed SFM points \cite{schoenberger2016sfm}.

We trained two variants of our model:
\begin{itemize}
\item\emph{Quark} starts from an initial LDM resolution of $64 \times 36$ and $24$ layers, and reduces the layers while increasing the spatial resolution to produce an output LDM of resolution $512 \times 288$ with $6$ layers. 

\item\emph{Quark+}, in contrast, starts from an initial LDM resolution of $96 \times 54$ and $32$ layers and produces an output LDM of resolution $768 \times 438$ with $8$ layers.
\end{itemize}

Both are targeted to a 1080p ($1920 \times 1080 $) image size, requiring a final upsampling factor (during Upsample \& Activate) of 3.75x and 2.4x for Quark and Quark+, respectively. See Tabs.~\ref{table:supp_quark} and~\ref{table:supp_quarkplus} in Appendix~\ref{sec:app_attention} for more details.

%% file: figures/table_comparison.tex
\begin{table*}[t]
  \centering
  \caption{Comparison to generalizable view synthesis methods. We evaluate all methods at 102 4x768 resolution and outperform the other baselines with faster runtime. Our \emph{Quark+} model offers a consistent quality boost over \emph{Quark}. Our model and ENeRF are also capable of generating at 2048x1536 resolution; we can outperform ENerf with real-time frame rates. Note that the \emph{TIME} column includes both reconstruction and rendering time. \lucy{if we don't like the 3 colors can search and remove cello and celly} %
  } 
  \label{table:comparison}

\begin{tabular}{ll|ccc|ccc|ccc|c}
\multicolumn{2}{c|}{Dataset} & \multicolumn{3}{c|}{Real Forward Facing (RFF)}  & \multicolumn{3}{c|}{NeX-Shiny Dataset} & \multicolumn{3}{c|}{Neural 3D Video} &  \\
Resolution & Method &  $PSNR^\uparrow$ & $SSIM^\uparrow$ & $LPIPS^\downarrow$  & $PSNR^\uparrow$ & $SSIM^\uparrow$ & $LPIPS^\downarrow$ & $PSNR^\uparrow$ & $SSIM^\uparrow$ & $LPIPS^\downarrow$ & $TIME^\downarrow$ \\
\hline

\multirow{ 5}{*}{1024x768} 
& SIMPLI & 23.35 & 0.812 & \cello 0.191 & 25.42 & \cello 0.844 & \cello 0.132 & 28.79 & 0.919 & \cello 0.169 & 4.4s\\
& GPNR & \cello 25.42 & \cello 0.816 & 0.218 & \cello 25.98 & 0.840 & 0.160 & 29.80 & 0.916 & 0.215 & 25s\\
& ENeRF & 23.71 & 0.780 & 0.224 & 25.75 & 0.836 & 0.143 & \cello 29.86 & \cello 0.922 & 0.172 & \cello 205ms \\
& \emph{Quark} & \celly 26.03 & \celly 0.850 & \celly 0.137 & \celly 26.51 & \celly 0.866 & \celly 0.108 & \celly 31.57 & \celly 0.942 & \celly 0.127 & \cellg 32.2ms \\
& \emph{Quark+} & \cellg 26.53 & \cellg 0.861 & \cellg 0.127 & \cellg 26.67 & \cellg 0.869 & \cellg 0.104 & \cellg 31.95 & \cellg 0.945 & \cellg 0.122 & \celly 91.9ms \\

\hline
\multirow{ 3}{*}{2048x1536}
& ENerf & \cello 23.62 & \cello 0.775 & \cello 0.308 & \cello 24.21 & \cello 0.780 & \cello 0.218 & \cello 29.62 & \cello 0.925 & \cello 0.246 & \cello 811ms \\
& \emph{Quark} & \celly 25.42 & \celly 0.809 & \celly 0.233 & \celly 24.66 & \celly 0.799 & \celly 0.174 & \celly 31.46 & \celly 0.940 & \celly 0.187 & \cellg 33.0ms \\
& \emph{Quark+} & \cellg 25.91 & \cellg 0.820 & \cellg 0.223 & \cellg 24.83 & \cellg 0.804 & \cellg 0.169 & \cellg 31.86 & \cellg 0.941 & \cellg 0.185  & \celly 92.6ms  \\

\bottomrule
\end{tabular}
\end{table*}

%% file: main_sections/05_results.tex
\section{Results}
\label{sec:results}
We provide qualitative and quantitative evaluations that demonstrate the capability of our method on a wide variety of static and dynamic scenes. We further perform a series of ablations to investigate the design choices involved in our model.

\input{figures/table_dl3dv.tex}
\input{figures/table_comparisons3dgs.tex}

\subsection{Comparisons} \label{sec:results_comparisons}

Below we compare our model to both generalizable (Tab.~\ref{table:comparison}) and non-generalizable (Tabs.~\ref{table:dl3dv} and~\ref{tab:comparisons3dgs}) view-synthesis approaches. Unless otherwise stated, we follow the standard evaluation procedure of using every eighth view as the target view, excluding all target views from the set of possible input views. We average first within a scene and then across scenes. 

Our model targets 1080p resolution, but the benchmarks below include various resolutions. To ensure a fair comparison we adjust the aspect ratio of the internal LDM dimensions within our model accordingly, retaining the same total area per layer and thus the same computational requirements as our 1080p model. Note that regardless of the required output resolution, Quark (and Quark+) produce an LDM with approximately the same number of pixels per layer, and rely on upsampling to produce the final output resolution.

\paragraph{Generalizable Methods.} In Tab.~\ref{table:comparison} we compare to SIMPLI~\cite{solovev2023}, eNeRF~\cite{lin2022efficient}, and GPNR~\cite{suhail2022generalizable}, which predict the image at the target viewpoint by processing nearby input images with generalizable networks (described in more detail in Sec.~\ref{sec:related}). We retrain all baseline models with the same camera selection strategy as ours on the RFF and IBRNet datasets. %

In the top of Tab.~\ref{table:comparison}, we compare all models at  resolution on the standard static datasets, Real Forward Facing (RFF)~\cite{mildenhall2020nerf} and Shiny~\cite{Wizadwongsa_2021_CVPR}, and the dynamic video dataset, Neural 3D Video~\cite{li2022neural}. For the video dataset, we follow the protocol in Li \etal~\shortcite{li2022neural} to first trim all videos to the same length (300 frames), and then evaluate every tenth frame using the zero-th camera as the target.

Our results consistently outperform the other generalizable methods on all three datasets and is the only method that is capable of running in \textit{real-time}, more than 6$\times$ faster than the nearest competitor, eNeRF. Our \emph{Quark+} model offers improved reconstruction over \emph{Quark}, owing to the increased resolution of the LDM, and is still fast enough to be interactive. eNeRF is the only other method that is capable of reconstruction and rendering at 2048x1536 resolution on the same hardware, which our \emph{Quark} model also outperforms with over $20\times$ faster runtime.

\paragraph{Non-generalizable methods.} Next we compare our model to various non-generalizable methods, for which there are multiple standard benchmarks with published results across many methods. First, in Tab.~\ref{table:dl3dv}, we compare using the DL3DV~\cite{ling2023dl3dv} benchmark. This benchmark contains diverse scenes captured from handheld mobile cameras and drone videos, with more complex camera trajectories than datasets captured using camera rigs or forward facing datasets, like those in Tab.~\ref{table:comparison}. 

Remarkably, we find that our method is also competitive with methods that are optimized independently for each scene. Despite not being trained on this dataset, our method surpasses the quality of the top performing methods, Zip-NeRF~\cite{barron2023zipnerf} and Gaussian Splatting (Tab.~\ref{table:dl3dv}). %

In Tab.~\ref{tab:comparisons3dgs}, we also evaluate on twelve scenes from MipNerf360~\cite{barron2021mipnerf}, Tanks and Temples~\cite{knapitsch2017tanks}, and Playroom from Deep Blending~\cite{hedman2018deep} used in ~\cite{kerbl_2023}. Our method ranks third on Mip-NeRF360 and first on Tanks and Temples, while outperforming Instant-NGP~\cite{mueller2022instant} and Plenoxels~\cite{Fridovich-Keil_2023_CVPR} on Deep Blending (ranking third on LPIPS). We note that a key difference between optimization-based approaches and ours is that our model renders the target view from only the nearest eight input cameras, rather than optimizing for the entire scene over all input cameras. For many of the benchmarks, we find that the target views are well covered by the eight nearest cameras measured by L2 distance of the camera centers, but for Deep Blending this assumption does not hold. Thus, we found that a modified camera selection heuristic which also incorporates the difference between camera viewing angles improves the results over using L2 distance alone, due to the variation in viewing angles of nearby cameras. Even with this modified heuristic, our method %
produces significant artifacts when there is insufficient coverage (e.g. when part of the target view is only visible from one or two distant cameras outside of the eight nearest). We do note however, that per-scene optimization methods also produce artifacts in these low coverage areas, but their failures tend to be more graceful, producing blurrier outputs, whereas the blending weights in our method can produce inaccurate hard edges. We also note that some of our method's failures could be reduced using more sophisticated, perhaps learned, view selection method.
\ryan{this paragraph could be tightened.}

\paragraph{Qualitative results.} We  show several examples of our rendered outputs in Figs.~\ref{fig:renders-generalizable},~\ref{fig:renders-dl3dv}, and~\ref{fig:renders-3dgs} that demonstrate examples of high quality reconstruction across a broad range of scenes. As differences between methods are more apparent in video than still images, especially when comparing Quark and Quark+, please also see our accompanying video.

We first compare to the generalizable algorithms in Fig.~\ref{fig:renders-generalizable}. eNeRF struggles with scenes with significant depth complexity. This is likely due to the fact that their algorithm uses a single-layer depthmap (albeit with varying thickness). Quark and Quark+ both appear to produce sharper edges around complex geometry than any of eNeRF, SIMPLI, or GPNR. Relative to SIMPLI, Quark's sharper results are likely aided by its use of IBR with blend weights. Relative to GPNR, it's possible that Quark's LDM layers track the surfaces and edges better than GPNR's Transformer-based approach for aligning patches along viewing rays.\ryan{is this too speculative?}

In Fig.~\ref{fig:renders-dl3dv}, we compare to several offline non-generalizable methods using the DL3DV benchmark. 
All results are quite high quality, but there are some noticeable differences.
The most noticeable are in reflective and refractive objects in the first 4 rows. We suspect that Quark benefits from using view selection to pick a small number of input views near to the novel view and being able to optimize the LDM for that view, whereas the other methods create one global geometry in a preprocess with an imperfect physical representation of these difficult view-dependant materials. Close inspection of the bottom two rows of the results for the NeRF-based methods (InstantNGP, MultiNeRF360, Nerfactor, and ZipNeRF) reveals some speckling artifacts which are a common side-effect of ray march rendering. The results for all of Quark, Quark+, and 3DGS look comparably sharp, but Quark reconstructs and renders in real-time while 3DGS requires an expensive offline optimization process to reconstruct the scene.

Finally, in Fig.~\ref{fig:renders-3dgs}, we compare to a different set of offline non-generalizable methods using the MipNeRF-360 and Tanks and Temples datasets. Plenoxels produces blurrier results than the other methods, likely due to its sparse 3D grid representation. Quark, Quark+, INGP-Big, M-NeRF360, and 3DGS-30k are all quite close in visual quality (with 3DGS-30k being perhaps slightly sharper than the rest). Quark both reconstructs and renders these scenes in real-time (and Quark+ at interactive rates), while all of the other methods require minutes to hours for scene reconstruction. 

We note that none of the metrics above measure temporal flicker, and as can be seen in the accompanying videos, the offline, full scene methods such as NeRF and 3DGS outperform Quark in this regard.  Quark switches between input images as the target cameras moves within the scene. This can cause noticeable flickering in the rendered image, particularly when there are large exposure differences in the input images.

\subsection{Ablations}\label{sec:results_ablations}

We performed several ablations to evaluate our model design choices. We used the Quark+ network as our baseline for ablations, however for expediency, we trained on half resolution training data and for only 250k training steps.
We performed the evaluation of the ablations on the DL3DV dataset, at 1080p resolution. 
Note that this resolution differs from our previous DL3DV benchmark (where scores for competing methods were only available at a lower resolution). However, we found that the distinctions between the various ablations were more apparent at this increased resolution.

We note that the differences in scores below are small. Many of the scenes in DL3DV are relatively simple and the simplified approaches can produce adequate results. Even within more complex scenes the differences are often confined to a small set of pixels, often near object boundaries, leading to relatively small changes in the metrics. However, when viewed side-by-side the differences are quite apparent, and we encourage the interested reader to view the ablation results in the supplemental material.

In table~\ref{tab:ablation} we show the \% increase or decrease in runtime vs the baseline. These numbers are approximate as the ablated methods have not been optimized. For example, \emph{No cross attention} was implemented by simply zero-ing out the attention keys.

\begin{table}[]
\centering
\caption{Ablations of our method across several key design decisions. Difference in runtime speed are expressed as a percentage. We show quality differences using PSNR, SSIM and LPIPS. \label{tab:ablation}}
\small
\scalebox{0.95}{
\begin{tabular}{lrrrr}

Ablation & \multicolumn{1}{l}{\begin{tabular}[c]{@{}l@{}}Runtime \\ change \%\end{tabular}} & \multicolumn{1}{l}{$PSNR^\uparrow$} & \multicolumn{1}{l}{$SSIM^\uparrow$} & \multicolumn{1}{l}{$LPIPS^\downarrow$} \\
\midrule
No layer collapse & +115.18 & 30.5635 & 0.9184 & 0.1531 \\
\textbf{Baseline} & 0 & 30.4071 & 0.9165 & 0.1543 \\
Fewer iterations & -17.49 & 30.3227 & 0.9153 & 0.1554 \\
No ray directional encoding & -1.28 & 30.3179 & 0.9157 & 0.1555 \\
No render-and-refine & -5.22 & 29.7106 & 0.9111 & 0.1599 \\
No cross attention & -3.09 & 29.7289 & 0.9097 & 0.1621 \\
RGB Output & -4.56 & 29.1779 & 0.8926 & 0.1788 \\
\bottomrule
\end{tabular}
}
\end{table}

We performed the following ablations.
\textit{No layer collapse}: We remove the layer collapse modules. This leads to a much (over 2$\times$) slower algorithm but does have improved quality vs the baseline.
\textit{Baseline}: The Quark+ model, trained at lower resolution.
\textit{Fewer Iterations}: We tested the effect of reducing the number of iterations by removing the first and second \figblock{Update \& Fuse} iterations. To preserve the output resolution of the LDM, we doubled the resolution of the initial LDM.
\textit{No ray directional encoding}: To simulate the effect of removing the ray directional encoding, we replaced the ray directional encoding with zeroes.
\textit{No cross attention}: To simulate the effect of removing attention, we replaced all keys within the attention stages of the network with zeroes. This corresponds to a simple mean across the image update feature maps. Note that we only zero-ed out the keys within the core solver, we retained the full keys when computing the blend weights.
\textit{No render-and-refine}: To simulate the effect of removing the render and refine stage of our network, we replaced the rendered image with zeroes.
\textit{RGB Output}: Instead of computing RGB through blending the input images, we output RGB directly from the network. Specifically, we use the first three channels of the appearance feature used inside the network as the output RGB. 

Overall, the ablations demonstrate the importance of each of our contributions, ray directional encoding, cross attention, render-and-refine, and our blend weight strategy. Intriguingly, the fewer iterations ablation performs almost as well as the baseline, with a 17\% runtime reduction, pointing to other trade-offs. More generally, our network has many hyper parameters and a full exploration of all of the design space was not possible. It's likely that there are other configurations that perform equally well as ours with equivalent or lower runtime. An automated search through the parameter sweep would be interesting future work.

\ryan{Has this part been updated?}
We also evaluate the performance of our model with different camera baseline distances on the spaces dataset, similar to \cite{flynn2019deepview}. Tab.~\ref{table:baseline_ablation} shows the results of our method on the Spaces dataset while varying the camera baseline distance. Following the analysis in \cite{flynn2019deepview}, we evaluate on three different baseline sizes: small, medium, and large. For this experiment, we train a model that takes an arbitrary number of input images from 4--16. Our model is robust to varying camera baselines, achieving similar performance across the three baseline settings.
This is in contrast to DeepView, which has highest performance on the small baseline dataset. Note that our SSIM metrics are not directly comparable because of slight differences in input preprocessing.

\begin{table}
\centering
\caption{Ablation on camera baseline distance, evaluated on the Spaces dataset~\cite{flynn2019deepview}. Our results show consistent performance across camera baseline distances.} 
\label{table:baseline_ablation}
\begin{tabular}{l|ccc}
& \multicolumn{3}{c}{Spaces Dataset} \\
Baseline (cm) & $PSNR^\uparrow$ & $SSIM^\uparrow$ & $LPIPS^\downarrow$\\
\hline
Small ($\sim$10) $\quad$ & 29.03 & 0.9215 & 0.1586 \\
Medium ($\sim$20) $\quad$ & 29.57 & 0.9258 & 0.1541 \\
Large ($\sim$30) $\quad$ & 29.43 & 0.9241 & 0.1541 \\
\bottomrule
\end{tabular}
\end{table}

\subsection{Limitations}
\label{sec:limitations}

Our method produces state-of-the-art quality at real-time rates across a broad range of scenes. However, there are several notable limitations that present opportunities for future work.

We do not enforce any temporal consistency between frames for video reconstructions; each frame is reconstructed independently from its neighbors. This produces some flickering artifacts in video reconstructions (see the accompanying video) and presents an opportunity to explore the temporal domain in order to produce smoother results.

We use blend weights and IBR for efficient and high-resolution rendering, but this also results in a few side-effects. Blend weights alone cannot accurately represent view dependent materials, as the per-pixel color produced from our model is a convex combination of the colors from the input images. Fortunately, our ability to optimize the blend weights for each frame, in particular using the ray direction encoding described in Sec.~\ref{sec:method-encode-input}, makes it possible to convincingly reproduce some reflective and refractive surfaces, as can be seen in our results on the Shiny dataset in Tab.~\ref{table:comparison}, in Fig.~\ref{fig:renders-dl3dv} and in our accompanying video. Nonetheless, others have shown benefits by giving the network the representational power to more accurately model view dependent appearance~\cite{Wizadwongsa_2021_CVPR, verbin2022refnerf}, and it would be worth trying this within our algorithm as future work. 

Blend weights also struggle in situations where images are taken with different exposures or focal lengths. As can be seen in the accompanying video, this can lead to visible edge artifacts and flickering. Moreover, when there are errors in input camera calibration, our method may produce incorrect edges rather than the more pleasing blurring produced by other methods that don't use IBR.

Finally, in order to achieve real-time rates, our method must use a small number of input images (e.g. 8--16). As demonstrated in Tab.~\ref{table:dl3dv}, this is effective when those input images consistently provide enough coverage to synthesize the novel view. However, this is not the case for some datasets (e.g. Mip-NeRF360 and Deep Blending in Tab.~\ref{tab:comparisons3dgs}). This problem is also visible in some of our accompanying video results, and tend to show up as artifacts near the edges of the novel view.  %
In these situations, a better view-selection algorithm or a method that optimizes a single geometry over all images may produce better results.

%% file: figures/table_dl3dv.tex
\begin{table}
\centering
\caption{Comparison to non-generalizable methods on the DL3DV-10K benchmark. Baseline results are directly obtained from~\citet{ling2023dl3dv}. Our model attains the best quality without any per-scene optimization.\lucy{I removed Zip-Nerf*, I don't think the distinction is relevant for us.}}
\label{table:dl3dv}
\begin{tabular}{l|ccc}
Dataset & \multicolumn{3}{c}{DL3DV-10K} \\
Method & $PSNR^\uparrow$ & $SSIM^\uparrow$ & $LPIPS^\downarrow$ \\
\hline
Instant-NGP              & 25.01            & 0.834     & 0.228             \\
Nerfacto                 & 24.61            & 0.848     & 0.211             \\
Mip-NeRF 360             & 30.98            & 0.911     & 0.132             \\
3DGS                     & 29.82            & 0.919     & 0.120             \\
Zip-NeRF                 & \cello 31.22            & \cello 0.921     & \cello 0.112    \\
\hline
\emph{Quark} &  \celly 31.31 & \celly 0.932 & \celly 0.103 \\
\emph{Quark+} & \cellg 31.53 & \cellg 0.935 & \cellg 0.101 \\
\bottomrule
\end{tabular}
\end{table}

%% file: figures/table_comparisons3dgs.tex
\begin{table*}[!ht]
    \centering
    \caption{Comparison to non-generalizable methods on the Gaussian Splatting datasets. Baseline results are directly obtained from~\citet{kerbl_2023}. Our method is competitive on Mip-NeRF360 and Tanks\&Temples. However, Quark is more greatly affected by the selection strategy for the limited number of input views (especially for Deep Blending) as opposed to the remaining methods which optimize each scene over all views. 
    \label{tab:comparisons3dgs} }
    \small
    \scalebox{1.15}{
    \begin{tabular}{l|ccc|ccc|ccc}
        Dataset & \multicolumn{3}{c|}{Mip-NeRF360}  & \multicolumn{3}{c|}{Tanks\&Temples} & \multicolumn{3}{c}{Deep Blending (Playroom)}\\
        Method
        & $PSNR^\uparrow$   & $SSIM^\uparrow$    & $LPIPS^\downarrow$  
        & $PSNR^\uparrow$   & $SSIM^\uparrow$    & $LPIPS^\downarrow$  
        & $PSNR^\uparrow$   & $SSIM^\uparrow$    & $LPIPS^\downarrow$   \\
        \hline
        Plenoxels & 23.08 & 0.626 & 0.463 
        & 21.08 & 0.719 & 0.379 
        & 23.02 & 0.802 & 0.418 \\
        INGP-Base & 25.30 & 0.671 & 0.371 
        & 21.72  & 0.723 & 0.330 
        & 19.48 & 0.754 & 0.465 \\
        INGP-Big & 25.59 & 0.699 & 0.331 
        & 21.92 & 0.745 & 0.305 
        & 21.67 & 0.780 & 0.428 \\
        M-NeRF360 & \cellg  27.69 &\celly 0.792 & 0.237 
        & 22.22 & 0.759 & 0.257 
        & \celly 29.66 & \celly 0.900 & \celly 0.252 \\
        GS-7K  & 25.60 & 0.770 & 0.279 
        & 21.20 & 0.767 & 0.280 
        & \cello 29.24 & \cello  0.896 & 0.291 \\
        GS-30K &  \celly 27.21 & \cellg 0.815 & \celly 0.214 
        & \cello 23.14 & \cello 0.841 & \cello 0.183 
        & \cellg 30.04 & \cellg 0.906 & \cellg 0.242 \\
        \hline
        
\emph{Quark} & 26.60 & 0.774 & \celly 0.214 
& \celly 23.60 & \celly 0.852 & \celly 0.138 
& 23.61 & 0.837 & 0.289 \\
\emph{Quark+} & \cello 26.91 & \cello 0.784 & \cellg 0.207 
& \cellg 23.83 & \cellg 0.857 & \cellg 0.133 
& 23.58 & 0.837 & \cello 0.287 \\

    \bottomrule
    \end{tabular}
}
\end{table*}

%% file: main_sections/06_conclusion.tex
\section{Conclusion}

We introduce a novel neural algorithm (Quark) for fast, high-resolution, and high-quality novel view synthesis. Our algorithm employs several optimizations to achieve real-time rates for combined scene reconstruction and rendering. Key to our approach is performing intermediate computations at low resolution -- we use an encoded LDM representation, which is iteratively rendered to the input views and refined via update and view fusion modules. Additional efficiency improvements come from an optimized One-to-many attention operation to incorporate information from multiple input views during view fusion, and layer collapse which reduces the number of layers as the spatial resolution of the intermediate LDM increases. We demonstrate state-of-the-art quality across a wide variety of test scenes. We outperform other generalizable view synthesis approaches on standard static and dynamic datasets, and either outperform or are competitive with even non-generalizable approaches that perform per-scene optimization. Altogether, our Quark model enables fully feed-forward novel view synthesis from a set of input cameras at up to 2K resolution, performing both scene reconstruction and rendering at real-time rates.

%% file: supp_sections/01_appendix.tex
\appendix
\section{One-to-Many Attention}
\label{sec:app_attention}

This section elaborates on the computational complexity of One-to-many attention. We follow the exposition from \cite{vaswani2017attention}, the notation used here should be considered independent from the rest of the paper.

\begin{table}[b]
\centering
\caption{Theoretical number of FLOPS vs the number of heads $h$ for both standard cross attention and our optimized One-to-many attention. Assumes $N=8$ inputs. The last column shows One-to-many attention's relative speed-up vs standard cross attention.}
\label{table:supp_one_to_many_versus_heads}
\small
\begin{tabular}{cccc}
\multicolumn{1}{p{1.5cm}}{\centering Number\\ of heads (h)} &
\multicolumn{1}{p{2.0cm}}{\centering Standard Cross\\ Attention (flops)} &
\multicolumn{1}{p{2.0cm}}{\centering One-to-Many\\ Attention (flops)} &
\multicolumn{1}{p{1.2cm}}{\centering Relative\\ Speed up} \\
\midrule
1 & 18688 & 2304 & 8.1x  \\
2 & 18688 & 4608 & 4.1x  \\
4 & 18688 & 9216 & 2.0x  \\
8 & 18688 & 18432 & 1.0x \\
\bottomrule
\end{tabular}
\end{table}

\begin{table}[b]
\centering
\caption{Theoretical number of FLOPS vs the number of inputs $N$ for both standard cross attention and our optimized One-to-many attention when using $h=4$ attention heads. The last column shows One-to-many attention's relative speed-up vs standard cross attention.}
\label{table:supp_one_to_many_versus_images}
\small
\scalebox{0.95}{
\begin{tabular}{ccccc}
\multicolumn{1}{p{1.3cm}}{\centering Number of\\inputs (N)} &
\multicolumn{1}{p{1.7cm}}{\centering Standard Cross\\Attention (flops)} &
\multicolumn{1}{p{1.1cm}}{\centering Relative\\Slow Down} &
\multicolumn{1}{p{1.6cm}}{\centering One-To-Many\\Attention (flops)} &
\multicolumn{1}{p{1.2cm}}{\centering Relative\\Slow Down} \\
\midrule
4  & 10368 & 1.0 & 8704 & 1.0x   \\
8  & 18688 & 1.8x & 9216 & 1.1x   \\
16 & 35328 & 3.4x & 10240 & 1.2x   \\
32 & 68608 & 6.6x & 12288 & 1.4x   \\
64 & 135168 & 13.0x & 16384 & 1.9x  \\
\bottomrule
\end{tabular}
}
\end{table}

Dot product attention computes a weighted sum over values $V$ by computing the \emph{softmax} of the dot product between query vectors $Q$ against key vectors $K$.

\begin{equation}
\text{Attention}(Q, K, V) = \text{softmax}\left(\frac{QK^T}{\sqrt{d_k}}\right)V
\end{equation}

In multi-head attention, the query, keys and values are all projected to $h$ heads before computing attention. That is,

\begin{equation}
\text{MultiHead}(Q, K, V) = \text{concat}(\text{head}_{1}, \dots, \text{head}_{h})W^O
\end{equation}

with the $i$-th head defined as

\begin{equation}
\begin{split}
\text{head}_i &= \text{Attention}\left(QW_i^Q, KW_i^K, VW_i^V\right)\\
&= \text{softmax}\left(\frac{QW_i^Q(W_i^K)^TK}{\sqrt{d^h}}\right)VW_i^V
\end{split}
\end{equation}

Here we note the redundancy: instead of applying both $W_i^Q$ and $W_i^V$ when computing $\text{head}_i$ we could instead fold $W_i^Q\left(W_i^K\right)^T$ into a single matrix and apply it to $Q$ alone. This typically does not provide a reduction in total operations since the number of channels, $d^h$, is also reduced for each head (typically to $d_k/h$), hence it is more efficient to project the queries and keys to a reduced number of channels before computing their dot product. Additionally, for \emph{self attention} the number of queries matches the number of keys so the savings from performing the matrix multiply only on  the queries is a factor of two.
\begin{figure}[t]
    \centering
    \includegraphics[width=\columnwidth]{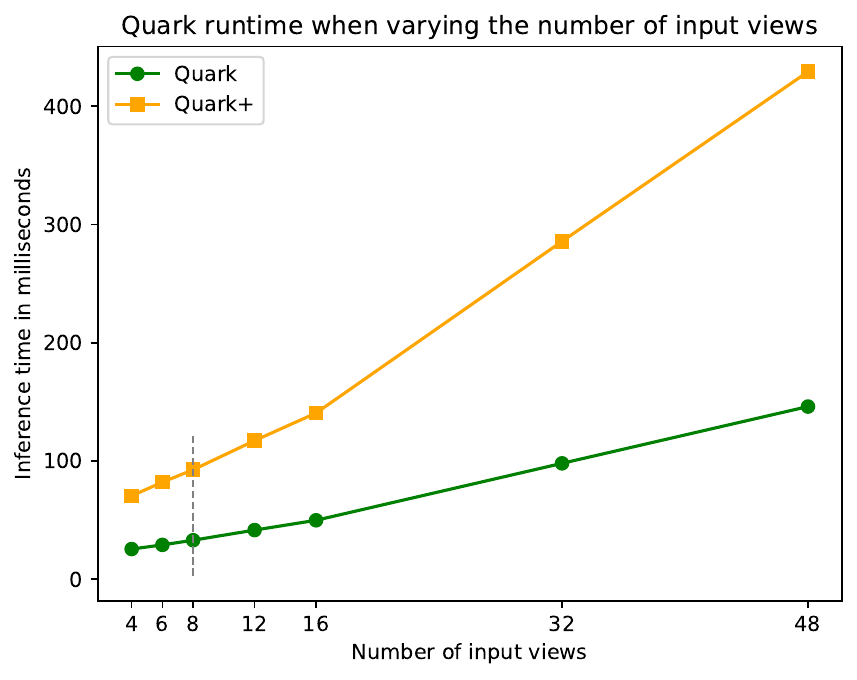}
    \caption{Quark inference time for different numbers of input views. Inference time scales linearly in the number of views and the Quark model achieves interactive (> 10 frames per second) frame rates up to 32 input views. }
    \label{fig:number_of_input_views}
\end{figure}

However for our cross-attention use case, we have a single query and a relatively small number of heads, a significant speed up can be achieved by folding the matrix into the query. 

Specifically, in standard cross-attention with a single query attending against N heads. This require $(N+1) \times h$ matrix multiplies, each of size $d_k \times \frac{d_k}{h}$ to extract the query and reference keys, and an additional $N \times h$ matrix multiplies of the same size to extract the values, leading to a total of $(2N+1) \times h \times d_k \times \frac{d_k}{h} = (2N+1) \times d_k^2$ floating point operations. To compute the $N \times h$ dot products we require $N \times h \times \frac{d_k}{h} = N \times d_k$ operations. Finally to compute the output we concatenate $h \times \frac{d_k}{h}$ and then perform a single $d_k \times d_k$ matrix multiply requiring $d_k^2$ operation. The total number of operations for standard 1-to-$N$ cross attention is thus:

\begin{equation}
    (2N+1) \times d_k^2 + N \times d_k + d_k^2
\end{equation}

In contrast, for the proposed one-to-many attention, we again have a single key attending against N heads, but now require $h$ matrix multiplies, each of size $d_k \times d_k$ to extract the $h$ query keys for a total of $h \times d_k^2$ floating point operations. To compute the $N \times h$ dot products we require $N \times h \times d_k$ operations. Finally to compute the output we concatenate $h \times d_k$ arrays and then perform a single $d_k \times (h \times d_k)$ matrix multiply requiring $h \times d_k^2$ operations. The total number of operations for one-to-many attention is thus:

\begin{equation}
    h \times d_k^2 + N \times h \times d_k + h \times d_k^2
\end{equation}
 
 Using the values from the method, $d_k=32$, $N=8$ and $h \in \{1,2,4\}$ we show the relative speed up in terms of flops below:

We include 8 heads here for completeness, but our algorithm never uses more than 4 heads, and at high resolutions it uses just 1 or 2 heads, leading to large savings, as shown in Table  \ref{table:supp_one_to_many_versus_heads}.

We can also examine the effect of varying the number of input keys, $N$, as $h$ is held constant, here at a value of 4. We show the relative slow down as we increase the number of inputs in Table \ref{table:supp_one_to_many_versus_images}. As can be seen, for standard cross attention, the number of floating point operations is approximately $O(N)$ but for one-to-many attention it is close to $O(1)$. 

Note that this analysis is only for the fusion block within our network, rendering to the input images and computing the image residuals is still an $O(\mathrm{number of images})$ operation, and hence the runtime increases with the number of input images as discussed below.

\begin{figure*}[t]
    \centering
    \includegraphics[width=\linewidth]{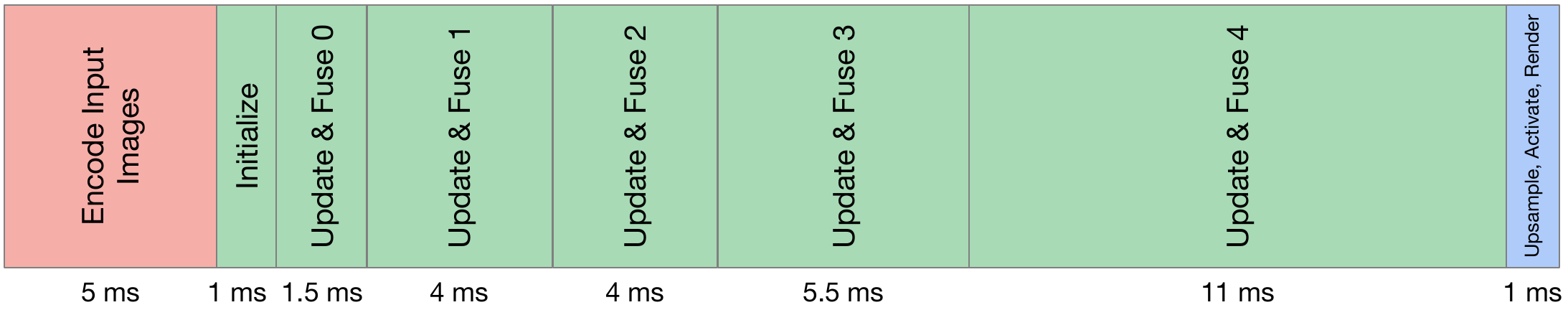}
    \caption{Timing diagram for the Quark algorithm running on an NVidia A100 GPU. Most of the inference time is spent performing iterative Update \& Fuse steps that iteratively refine the layered depth map. The final upsample \& activation as well as rendering is not a significant portion of the network run-time. }
    \label{fig:timing}
\end{figure*}

\section{Performance}
The run-time of our method is approximately $T_V + M*T_{image}$, i.e. there's a fixed base cost $T_V$ associated with operations on the encoded feature volume and rendering, and an additional a per-image cost. The base cost is dominated by the ~$O(1)$ One-to-many complexity plus the cost of the CNN applied to $V$ within the Fusion blocks.  The per image cost is dominated by image feature encoding and image update computation, which are both $O(M)$ operations. A detailed timing breakdown is shown in Fig.~\ref{fig:timing}, which shows the time for Quark to render a single frame, broken down by the sub-stages of the network shown in Fig.~\ref{fig:teaser}. 
The change in run-time versus number of input image is shown in Fig.~\ref{fig:number_of_input_views}. 

\section{Network details}
\begin{figure*}[t]
    \centering
    \includegraphics[width=\linewidth]{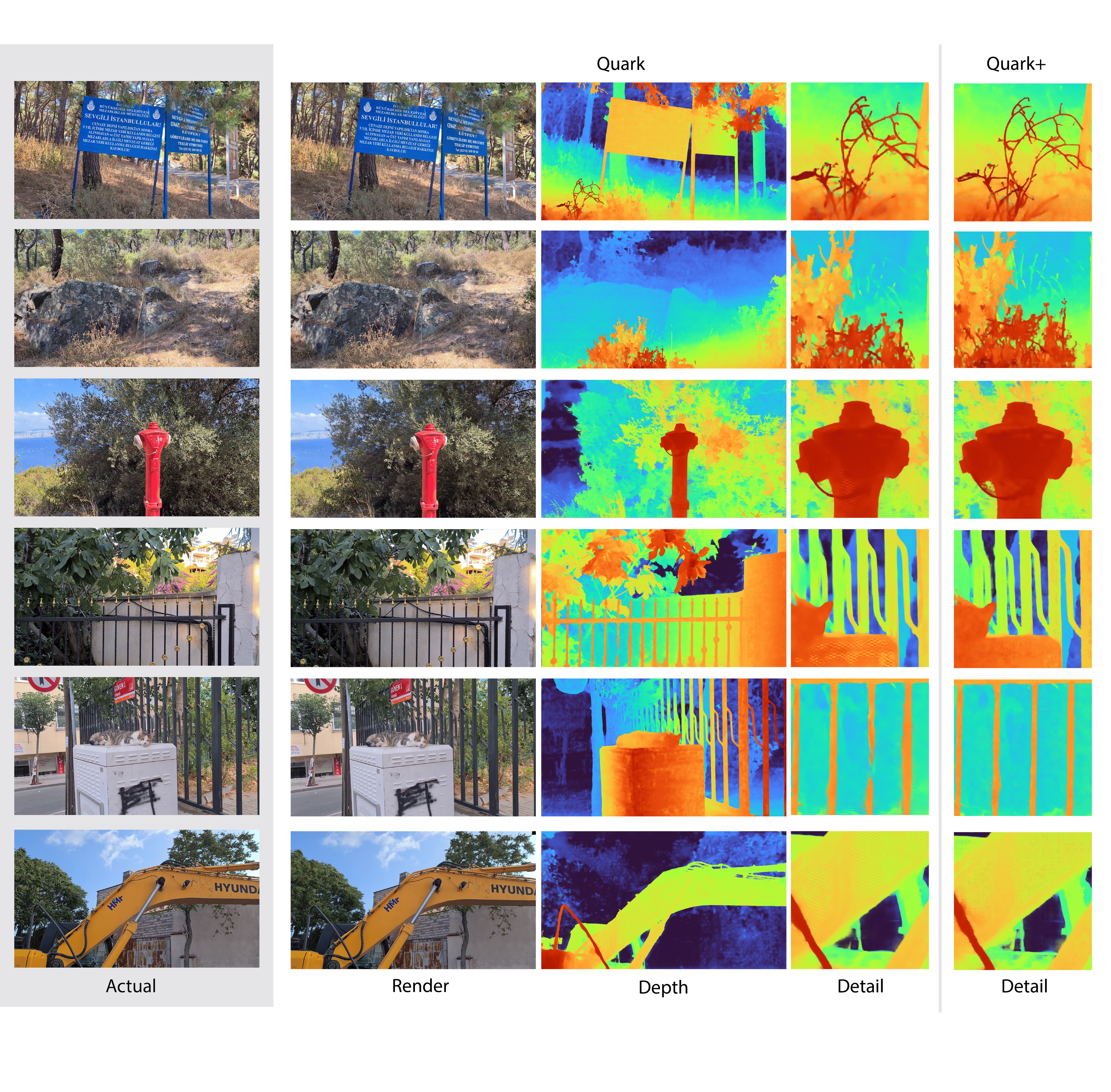}
    \caption{Quark and Quark+ renders and depth maps on scenes from the SWORD dataset.}
    \label{fig:renders_sword}
\end{figure*}

Tables ~\ref{table:supp_quark} and~\ref{table:supp_quarkplus} show the full details for the network components for Quark and Quark+ respectively. The increased dimensions of the \figblock{Update \& Fuse} steps for Quark+ helps Quark+ better capture fine details. This is demonstrated in Fig.~\ref{fig:renders_sword}, which compares Quark against Quark+ using the SWORD dataset. The differences are easiest to see in the depth map close-ups on the right.

\begin{table*}[b]
\small
\scalebox{1.1}{
\begin{tabular}{l|l|l|l|l}
Step & \begin{tabular}[c]{@{}l@{}}Input layer dimensions\\ (\#num layers, h, w, C)\end{tabular} & \begin{tabular}[c]{@{}l@{}}Encoded image \\ feature map dimensions\\ (h, w, C)\end{tabular} & Blocks & \begin{tabular}[c]{@{}l@{}}Output layer dimensions\\ (\#num layers, h, w, C)\end{tabular}  \\
\midrule
Initialize & \begin{tabular}[c]{@{}l@{}}1, 1, 1, 32 \\ (single broadcast
feature)\end{tabular} & 36, 60, 32 & Bp, A4, C, C, A4, C, C, A4, C, C & 24, 36, 64, 32  \\
\hline
Update \& Fuse 0 & 24, 36, 64, 32 & 36, 60, 32 & U, A4, C, C, A4, C, C, A4, C, C & 24, 36, 64, 32  \\
Update \& Fuse 1 & 24, 36, 64, 32 & 72, 120, 32 & U, A4, C, C, A4, C, C, A4, C, C & 24, 72, 128, 32 \\
Update \& Fuse 2 & 24, 72, 128, 32 & 72, 120, 32 & U, A4, C, A4, C & 24, 72, 128, 32 \\
Update \& Fuse 3 & 24, 72, 128, 32 & 144, 240, 32 & Lc, U, A2, C, A2, C & 12, 144, 256, 32 \\
Update \& Fuse 4 & 12, 144, 256, 32 & 288, 480, 32 & Lc, U, A1, C, A1, C & 6, 288, 512, 32  \\
\hline
Output & 6, 288, 512, 32 & 1080, 1920, 3 & \begin{tabular}[c]{@{}l@{}}Upsample,\\Decode density, depth and blend weights,\\Resample images and blend \end{tabular} & 8,1080, 1920, 4 (RGB + $\mathbf{\sigma}$) \\
\bottomrule
\end{tabular}
}
\caption{Model description for \emph{Quark}. The \figblock{Initialize} step starts with back-projecting the encoded images onto flat LDM layers, indicated by \textbf{Bp} and initializes the encoded LDM $V$ to a single 32-channel broadcasted feature. Each \figblock{Update \& Fuse} step starts with rendering and computing image update features and back-projecting them to the LDM, indicated by \textbf{U}. Within both the \figblock{Initialize} and {Update \& Fuse} steps we have several fusion blocks, shown here expanded into their constituent \figblock{One-to-many attention} and residual CNN blocks. Each \figblock{Fusion Block} contains a single \figblock{One-to-many Attention} block with $h$ heads, indicated by \textbf{A{h}}, followed by one or more residual CNN blocks, indicated by \textbf{C}. The final two \figblock{Update \& Fuse} steps start with a (\textbf{Lc}) \figblock{Layer Collapse} block that reduce the number of layers by 2x. This layer collapse occurs prior to any other blocks.}

\label{table:supp_quark}
\end{table*}

\begin{table*}[b]
\small
\scalebox{1.1}{
\begin{tabular}{l|l|l|l|l}
Step & \begin{tabular}[c]{@{}l@{}}Input layer dimensions\\ (\#num layers, h, w, C)\end{tabular} & \begin{tabular}[c]{@{}l@{}}Encoded image \\ feature map dimensions\\ (h, w, C)\end{tabular} & Blocks & \begin{tabular}[c]{@{}l@{}}Output layer dimensions\\ (\#num layers, h, w, C)\end{tabular}  \\
\midrule
Initialize & \begin{tabular}[c]{@{}l@{}}1, 1, 1, 32 \\ (single broadcast
feature)\end{tabular} & 54, 90, 32 & Bp, A4, C, C, A4, C, C, A4, C, C & 32, 54, 96, 32  \\
\hline
Update \& Fuse 0 & 32, 54, 96, 32 & 54, 90, 32 & U, A4, C, C, A4, C, C, A4, C, C & 32, 54, 96, 32  \\
Update \& Fuse 1 & 32, 54, 96, 32 & 108, 180, 32 & U, A4, C, C, A4, C, C, A4, C, C & 32, 108, 192, 32 \\
Update \& Fuse 2 & 32, 108, 192, 32 & 108, 180, 32 & U, A4, C, A4, C & 32, 108, 192, 32 \\
Update \& Fuse 3 & 32, 108, 192, 32 & 216, 360, 32 & Lc, U, A2, C, A2, C & 16, 216, 384, 32 \\
Update \& Fuse 4 & 16, 216, 384, 32 & 432, 720, 32 & Lc, U, A1, C, A1, C & 8, 432, 768, 32  \\
\hline
Output & 8, 432, 768, 32 & 1080, 1920, 3 & \begin{tabular}[c]{@{}l@{}}Upsample,\\Decode density, depth and blend weights,\\Resample images and blend \end{tabular} & 8,1080, 1920, 4 (RGB + $\mathbf{\sigma}$) \\
\bottomrule
\end{tabular}
}
\caption{Model description for \emph{Quark+}. See caption of Tab.~\ref{table:supp_quark} for notation shorthand.}\label{table:supp_quarkplus}
\end{table*}

